\documentclass{article}
\usepackage{PRIMEarxiv}
\usepackage[utf8]{inputenc}
\usepackage[T1]{fontenc}
\usepackage{hyperref}
\usepackage{url}
\usepackage{booktabs}
\usepackage{amsfonts} 
\usepackage{amsmath}
\usepackage{nicefrac}     
\usepackage{microtype}    
\usepackage{lipsum}
\usepackage{graphicx}
\usepackage{xcolor}
\usepackage{array}
\usepackage{longtable}
\hypersetup{
	colorlinks=false,
	linkbordercolor=white,
	urlbordercolor=white,
	pdfborder={0 0 0}
}

\title{Transformer-based Spatial Grounding: A Comprehensive Survey}

\author{
	Ijazul Haq \\
	Shien-Ming Wu School of Intelligent Manufacturing, \\
	South China University of Technology, \\
	Guangzhou, 511442, China.\\
	Zirak.ai \\
	\And
	Muhammad Saqib \\
	Department of Software Engineering, \\
        University of Engineering \& Technology, \\
        25000, Peshawar, Pakistan \\
        \And
	Yingjie Zhang \\
	Shien-Ming Wu School of Intelligent Manufacturing, \\
	South China University of Technology, \\
	Guangzhou, 511442, China \\
}

\begin{document}
\maketitle

\begin{abstract}
Spatial grounding, the process of associating natural language expressions with corresponding image regions, has rapidly advanced due to the introduction of transformer-based models, significantly enhancing multimodal representation and cross-modal alignment. Despite this progress, the field lacks a comprehensive synthesis of current methodologies, dataset usage, evaluation metrics, and industrial applicability. This paper presents a systematic literature review of transformer-based spatial grounding approaches from 2018 to 2025. Our analysis identifies dominant model architectures, prevalent datasets, and widely adopted evaluation metrics, alongside highlighting key methodological trends and best practices. This study provides essential insights and structured guidance for researchers and practitioners, facilitating the development of robust, reliable, and industry-ready transformer-based spatial grounding models.
\end{abstract}


\section{Introduction}
\label{introduction}
Spatial grounding in images involves associating linguistic expressions, such as phrases or sentences, with corresponding regions or objects within visual scenes. This interdisciplinary area combines natural language processing, computer vision, and deep learning, particularly leveraging recent advances in transformer-based architectures. Transformers, initially proposed for sequence-to-sequence tasks in natural language processing \cite{jiang2019tiger}, have demonstrated remarkable performance gains across diverse applications, including image classification, object detection, and multimodal grounding \cite{lu2019vilbert}\cite{salgotra2024survey}. The rapid adoption and effectiveness of transformers in multimodal domains have spurred significant research attention toward their utility specifically in spatial grounding tasks, making it essential to systematically analyze and synthesize existing knowledge to guide future research directions.
Despite considerable advancements, the spatial grounding domain faces several challenges, primarily stemming from inherent complexities in integrating visual and textual modalities \cite{liu2024grounding}\cite{chen2022multi}. A major challenge lies in accurately aligning nuanced textual descriptions with precise image regions, often complicated by ambiguous language and visual clutter \cite{silberer2018grounding}. Additionally, selecting appropriate datasets and evaluation metrics to reliably benchmark transformer-based spatial grounding models remains problematic, with diverse standards leading to fragmented practices across studies \cite{koh2023grounding}. Consequently, comprehensive insights into methodological trends, dataset selection rationale, and metric consistency are critical for enhancing model development and evaluation rigor.
Motivated by these challenges and the significant potential of transformers in addressing spatial grounding complexities, we conducted a Systematic Literature Review (SLR) examining transformer-based spatial grounding research published between 2018 and 2025. The specific objectives of our survey are to:
\begin{itemize}
    \item Identify transformer-based model architectures employed for spatial grounding in images
    \item Determine datasets and benchmarks predominantly used for evaluating these models.
    \item Document evaluation metrics applied to assess the performance of transformer models on spatial grounding tasks.
    \item Uncover key trends, prominent practices, and methodological advancements through systematic analysis, specifically regarding publication patterns, dataset utilization, prevalent model architectures, evaluation metric evolution, and comparative benchmark performance.
    \item Evaluate the industrial applicability of transformer-based spatial grounding models by analyzing their domain coverage, dataset usage, performance metrics, and overall readiness for real-world deployment 
\end{itemize}
To systematically address these objectives, our research first extracted comprehensive data on the datasets commonly employed for evaluating spatial grounding, detailing their scale, annotation types, and image counts. Understanding dataset characteristics provides critical insights into their suitability for diverse grounding tasks, ranging from general-purpose object grounding to domain-specific contexts like remote sensing \cite{jain2022bottom}. Secondly, we identified and categorized transformer-based model architectures, noting their technical foundations, backbone integrations, and reported innovations. Examining these models reveals prevailing strategies, influential designs, and emerging methodologies guiding transformer utilization in visual grounding \cite{yeh2018unsupervised}\cite{kiros2018illustrative}. Thirdly, we documented all benchmarks and evaluation metrics utilized across these studies to assess model effectiveness and reliability, highlighting the variability and consistency in current evaluative practices \cite{li2023transformer}.
Beyond identifying commonly used resources and methods, our survey incorporates an in-depth statistical analysis to elucidate several important trends. Specifically, we analyzed the publication trend in transformer-based grounding research, illustrating the growth trajectory and highlighting the increasing academic interest over time. Furthermore, we evaluated dataset usage to pinpoint which datasets dominate the research landscape and assessed dataset scale to identify those most significant in terms of instances and annotations. We also examined the relationship between datasets and their adopted Region-of-Interest (ROI) formats, revealing insights into annotation practices and their implications for model training. Additionally, our statistical investigation identified prominent transformer architectures, emphasizing models that have substantially influenced the grounding domain. Lastly, we analyzed evaluation metrics trends, determining frequently adopted metrics, their comparative performance, and how they evolved temporally, thus offering comprehensive insights into methodological advancements and evaluative standards in transformer-based spatial grounding.
This survey contributes significantly to the spatial grounding literature by synthesizing fragmented research insights into a coherent overview. By systematically reviewing existing studies, categorizing transformer architectures, and analyzing evaluation practices, our survey clarifies prevailing standards, highlights methodological gaps, and provides a structured framework for future research. This comprehensive synthesis is beneficial for researchers and practitioners aiming to design, evaluate, and advance transformer-based models for spatial grounding in images, ultimately supporting robust and reliable model deployment in real-world multimodal applications.

\section{Background}
Spatial grounding in images refers to the task of linking natural language expressions to corresponding visual elements within an image. This process involves understanding both the semantic meaning of the language and the spatial or visual structure of the image. Traditionally, spatial grounding has been explored under various formulations such as Referring Expression Comprehension (REC), Phrase Localization, and Referring Expression Segmentation. These tasks require models to interpret linguistic cues and map them onto the correct regions or objects, which is inherently challenging due to ambiguity in language, the variability of visual scenes, and the need for cross-modal reasoning.
The foundation of early spatial grounding systems rested on Convolutional Neural Networks (CNNs) for visual feature extraction and Recurrent Neural Networks (RNNs) for language modeling. While these methods laid the groundwork for integrating vision and language, they often suffered from limitations in capturing long-range dependencies and complex interactions between modalities. The advent of attention mechanisms introduced a shift in this paradigm, culminating in the development of transformer architectures, which revolutionized both natural language processing and computer vision domains.
Transformers rely on self-attention mechanisms to capture dependencies across input sequences without relying on recurrence. Their encoder-decoder structure enables effective modeling of contextual information in language, while their scalability and parallelizability have made them suitable for large-scale learning tasks. Vision Tranformers (ViTs) extended this architecture to the visual domain by treating image patches as tokens and modeling their interactions similarly to words in text \cite{li2023transformer}. This architectural innovation opened the door to unified models capable of processing and fusing multimodal data, making transformers an ideal candidate for spatial grounding tasks.
The transition from CNN-based architectures to transformers in computer vision has yielded significant improvements in performance, particularly in tasks requiring fine-grained localization and cross-modal alignment. Models like DETR (Detection Transformer) \cite{li2025tacmt}, MDETR (Modulated DETR) \cite{su2019vl}, and TransVG \cite{deng2021transvg} have demonstrated state-of-the-art results in object detection and spatial grounding by combining visual backbones with transformer encoders and decoders. These models effectively align image and text embeddings, enabling precise region prediction based on natural language descriptions. Moreover, the use of pretrained language models (e.g., BERT, RoBERTa) and vision models (e.g., ViT, ResNet) as backbones further boosts the representational capacity of transformer-based grounding systems.
Spatial grounding has found applications across a range of domains. In natural images, tasks like REC are extensively studied using datasets such as RefCOCO, RefCOCO+, and RefCOCOg. In remote sensing, spatial grounding helps localize objects using spatial and semantic queries in satellite imagery, with benchmarks like RSVG and DIOR-RSVG gaining traction. Dialogue grounding, another emerging area, involves resolving referring expressions in multi-turn conversations and visual dialogues, requiring models to handle coreference and contextual shifts. These diverse domains present unique challenges, but all benefit from the multimodal reasoning capabilities inherent in transformer architectures.
Several surveys have touched on vision-language tasks or transformer architectures, but few focus explicitly on their intersection in spatial grounding. Existing reviews often provide high-level overviews of multimodal transformers, visual question answering, or object detection, without systematically addressing the nuances of grounding tasks  \cite{xiao2024towards}\cite{hendricks2018grounding}\cite{chen2022multi}. To the best of our knowledge, this work represents the first comprehensive survey dedicated to evaluating transformer-based models specifically for spatial grounding in images. Our SLR bridges this gap by offering a focused, data-driven synthesis of model architectures, datasets, benchmarks, and evaluation methodologies used in this field.

\section{Research Methodology}
\begin{figure}[ht]
\setlength{\itemsep}{0pt}
    \centering
    \includegraphics[width=0.9\linewidth]{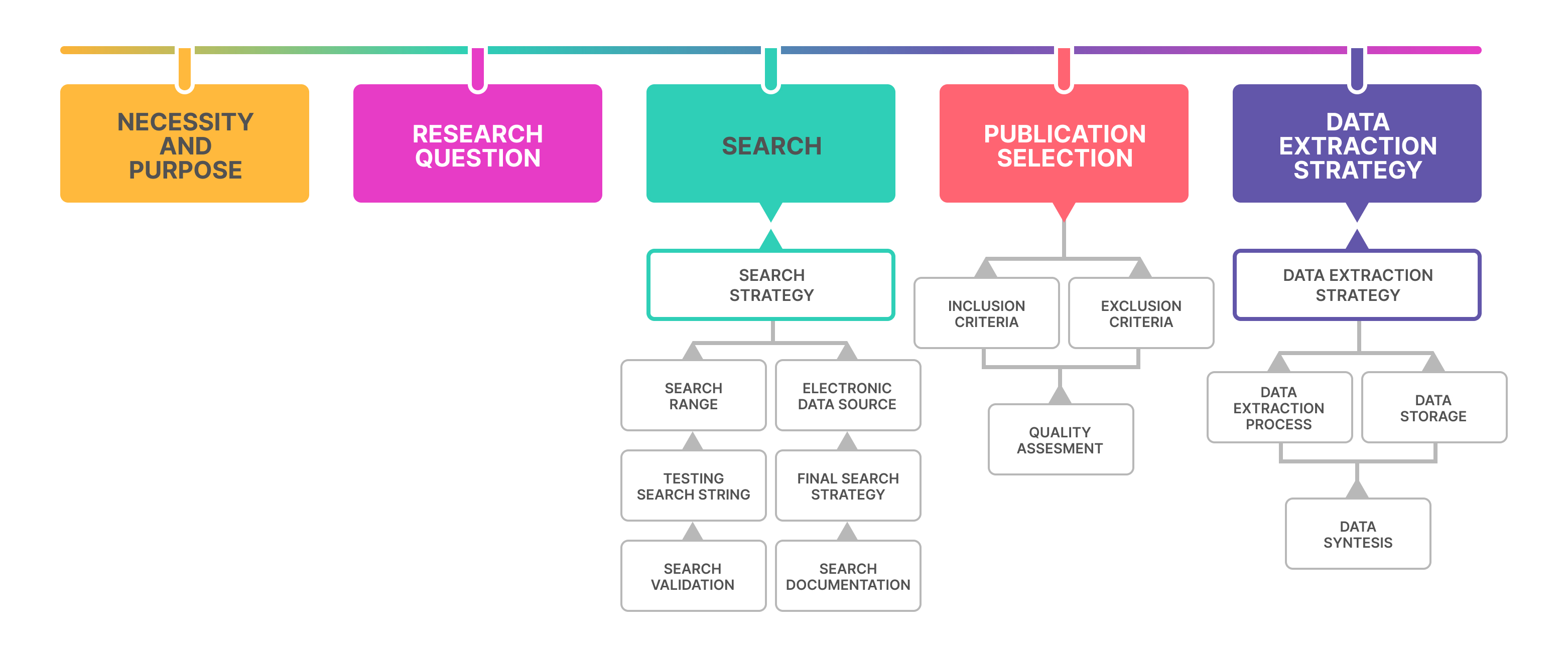}
    \caption{SLR Process}
    \label{figure_1}
\end{figure}

To ensure a structured, transparent, and reproducible review of existing literature on transformer-based models for grounding tasks, we adopted the principles of a SLR as outlined by Kitchenham et al \cite{kitchenham2009systematic}. The primary goal of this methodology is to comprehensively identify, evaluate, and synthesize all relevant studies that investigate the usage of transformer architectures for grounding across various modalities and applications. This methodological approach is particularly valuable in emerging and multidisciplinary fields like grounded multimodal AI, where research is dispersed across domains such as natural language processing, computer vision, and robotics. The process involved defining focused research questions, establishing clear inclusion and exclusion criteria, conducting a thorough search across major scientific databases, and systematically extracting and analyzing key data from selected studies as shown in Figure \ref{figure_1}. This methodology enables us to identify current trends, common practices, research gaps, and future directions within this rapidly evolving area.

\subsection{Systematic Literature Review}
\subsubsection{Formulated Research Questions}
To systematically investigate the application of transformer-based models in spatial grounding tasks, we formulated the following research questions (RQs), each addressing a specific yet interconnected dimension of the problem space. These questions guided our search strategy, informed our inclusion and exclusion criteria, and shaped our data extraction process:
\begin{itemize}
    \item \textbf{RQ1:} What transformer-based model architectures have been employed for spatial grounding in images?
    \item \textbf{RQ2:} Which datasets and benchmarks are predominantly used to evaluate these models?
    \item \textbf{RQ3:} What evaluation metrics are applied to assess the performance of transformer models on spatial grounding tasks?
    \item \textbf{RQ4:} What are the key trends, prominent practices, and methodological advancements identified through the systematic analysis of transformer-based spatial grounding research, specifically regarding publication patterns, dataset utilization, prevalent model architectures, evaluation metric evolution, and comparative benchmark performance?
    \item \textbf{RQ5:} How effectively do transformer-based spatial grounding models align with industrial application needs in terms of dataset utilization, performance reliability, and evaluation consistency?
\end{itemize}

\subsubsection{Construction of Search Terms}
We constructed a targeted search strategy using the Population, Intervention, Comparison, and Outcome (PICO) framework to align with the core components of our RQs. We identified key terms and synonyms for each element and used them to develop a comprehensive Boolean query, as shown in Table \ref{table_1}. We used this query to search five major academic databases: IEEE Xplore, ACM Digital Library, Springer Link, Elsevier, and Google Scholar.

\begin{table}[ht]
    \centering
    \caption{Search Terms for Transformer-Based Spatial Grounding}
    \label{table_1}
    \renewcommand{\arraystretch}{1.2}
    \begin{tabular}{|m{0.18\linewidth}|m{0.77\linewidth}|}
        \hline
        Components & Keywords \\
        \hline
        Population & 
        “transformer”, “BERT”, “Vision Transformer”, “ViT”, “LLM”, “CLIP”, “DETR”, “VLM”, “LVM”, “LMM” \\
        \hline
        Intervention & 
        “spatial grounding”, “visual grounding”, “language grounding”, “referring expressions” \\
        \hline
        Outcome & 
        Not explicitly defined; focus on analyzing model trends, datasets, evaluation metrics \\
        \hline
        Experimental Design & 
        “case study”, “experiment”, “empirical study”, “systematic literature review” \\
        \hline
        Boolean Query & (“transformer” OR “BERT” OR “ViT” OR “LLM” OR “CLIP” OR “DETR” OR “VLM” OR “LVM” OR “LMM”) AND (“spatial grounding” OR “visual grounding” OR “language grounding” OR “referring expressions”) AND (“case study” OR “experiment” OR “empirical study” OR “systematic literature review”) \\
        \hline
    \end{tabular}
\end{table}

\subsubsection{Publicaiton Selection and Data Extraction}
\begin{figure}[ht]
\setlength{\itemsep}{0pt}
    \centering
    \includegraphics[width=0.9\linewidth]{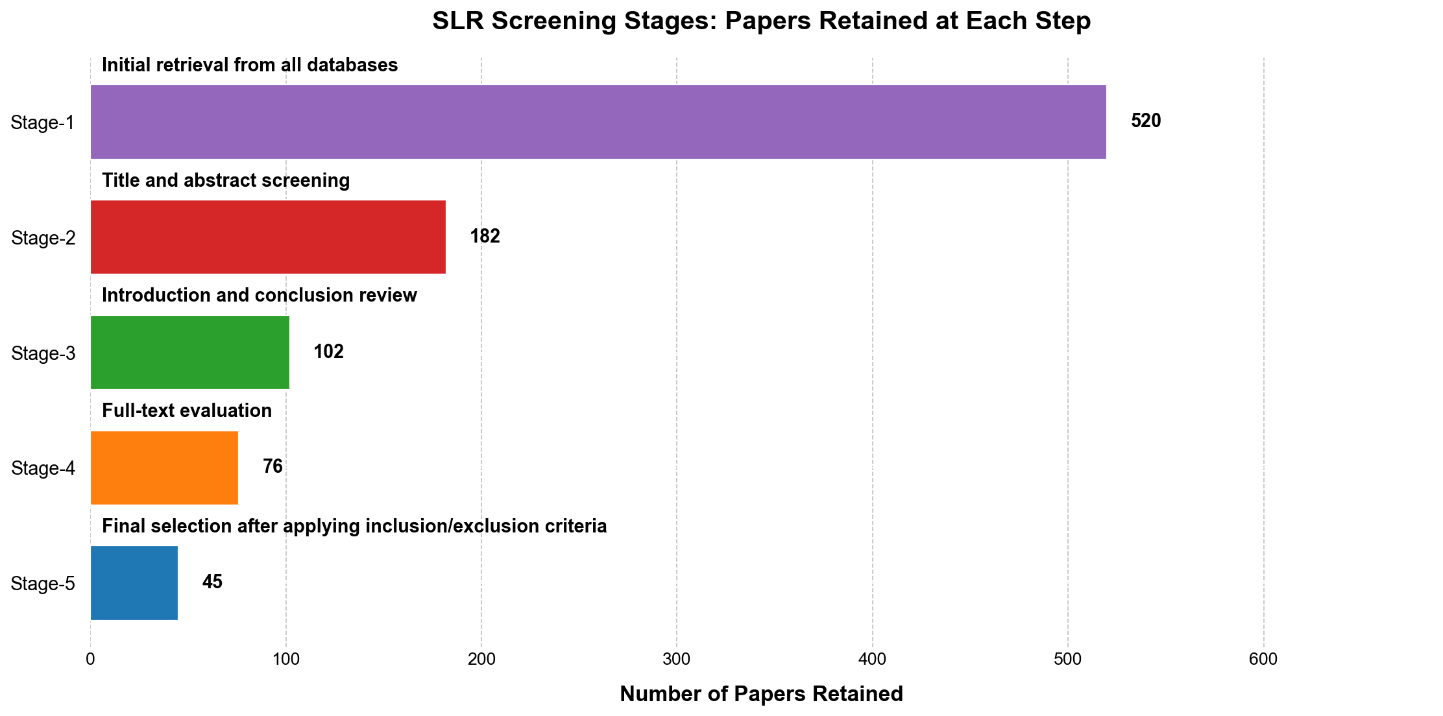}
    \caption{Publications Screening Process (Tollgate Approach)}
    \label{figure_2}
\end{figure}

In the initial retrieval phase, a total of 523 research articles were collected from major scientific databases using the predefined search string outlined in Section 3.1.2. To ensure methodological rigor, we applied the Tollgate approach \cite{afzal2009systematic} for refining the retrieved studies across multiple stages, including title/abstract screening, introduction/conclusion review, and full-text analysis based on predefined inclusion and exclusion criteria. Through this multistage filtering process, we identified 60 primary studies from 2018 to 2025 that met the quality standards and relevance criteria for our systematic review. The detailed shortlisting process is summarized in Table \ref{table_2} and Figure \ref{figure_2}.

\begin{table}[ht]
    \centering
    \caption{Selection of publications using Tollgate method}
    \label{table_2}
    \renewcommand{\arraystretch}{1.2}
    \begin{tabular}{|l|c|c|c|c|c|}
        \hline
        Digital Library       & Stage-1 & Stage-2 & Stage-3 & Stage-4 & Stage-5 \\ 
        \hline
        IEEE Xplore           & 80      & 38      & 32      & 18      & 5       \\ 
        ACM Digital Library   & 120     & 45      & 22      & 12      & 6       \\ 
        Springer              & 63      & 13      & 19      & 9       & 5       \\ 
        Elsevier              & 100     & 28      & 15      & 12      & 10      \\ 
        Google Scholar        & 160     & 58      & 14      & 25      & 19      \\ 
        \hline
        Total                 & 523     & 182     & 102     & 76      & 45      \\ 
        \hline
    \end{tabular}
\end{table}
From the 45 primary studies selected, we performed a structured data extraction to enable both qualitative interpretation and quantitative comparison. Our extraction process focused on six key dimensions: the architectural model type, the specific grounding task being addressed, the datasets used for training and evaluation, the metrics adopted for performance assessment, the core techniques applied, and the reported outcomes. These elements were carefully recorded for each study to uncover prevailing methodologies, evaluate effectiveness, and identify research gaps. A summary of the data extraction schema is presented in Table \ref{table_3}.

\begin{table}[ht]
    \centering
    \caption{Data Extraction Form}
    \label{table_3}
    \renewcommand{\arraystretch}{1.2}
    \begin{tabular}{|m{0.20\linewidth}|m{0.75\linewidth}|}
        \hline
        Dimension & Focus Area \\
        \hline
        Model Architecture & Transformer types used (e.g., BERT, LLM, VLM, ViT, CLIP, VisualGPT, DETR) \\
        \hline
        Grounding Task & Nature of the spatial grounding task (e.g., referring expression comprehension, object localization) \\
        \hline
        Datasets & Public benchmarks (e.g., RefCOCO, Flickr30K, Visual Genome) \\
        \hline
        Evaluation Metrics & Accuracy@1, IoU, Recall@K, mAP, precision/recall, grounding accuracy \\
        \hline
        Results & Key findings, ablation results, and reported trends \\
        \hline
    \end{tabular}
\end{table}

\subsubsection{Defining Inclusion and Exclusion Criteria}
We have defined the following inclusion and exclusion criteria as presented in Table \ref{table_4} and Table \ref{table_5}, with appropriate justifications to ensure the selection of the most relevant, high-quality research articles that effectively address our research questions concerning the use of transformer models for grounding tasks. This filtering framework guarantees that the final corpus of studies contributes meaningfully to the synthesis of datasets, techniques, and evaluation metrics in this domain.

\begin{table}[ht]
    \small
    \centering
    \caption{Inclusion Criteria}
    \label{table_4}
    \renewcommand{\arraystretch}{1.2}
    \resizebox{\textwidth}{!}{%
      \begin{tabular}{|c|m{8cm}|m{7cm}|}
        \hline
        Criterion & Description & Justification \\
        \hline
        I1 & Peer-reviewed journal articles, conference papers, or systematic reviews
           & Ensures scientific rigor and replicability through formal peer review. \\
        \hline
        I2 & Studies employing transformer-based architectures (e.g., BERT, ViT, CLIP, LLMs)
           & Central to the study’s focus on the role of transformers in grounding. \\
        \hline
        I3 & Research addressing visual or language grounding tasks, including referring expressions
           & Guarantees conceptual alignment with spatial grounding challenges. \\
        \hline
        I4 & Papers reporting empirical evaluations using standard datasets or benchmarks
           & Enables performance comparison and insight into generalizability. \\
        \hline
        I5 & Articles indexed in IEEE, ACM, Springer, Elsevier, or Google Scholar
           & Ensures accessibility, credibility, and coverage of scholarly sources. \\
        \hline
        I6 & Published between 2018 and 2025
           & Captures studies from the post-transformer revolution. \\
        \hline
        I7 & Written in English
           & Ensures accurate interpretation, consistency in analysis, and team-wide accessibility. \\
        \hline
      \end{tabular}%
    }
\end{table}

\begin{table}[ht]
    \small
    \centering
    \caption{Exclusion Criteria}
    \label{table_5}
    \resizebox{\textwidth}{!}{%
      \begin{tabular}{|c|m{8cm}|m{7cm}|}
        \hline
        Criterion & Description & Justification \\
        \hline
        E1 & Studies not involving transformer-based architectures
           & Irrelevant to the study’s primary interest in transformers. \\
        \hline
        E2 & Studies not addressing visual/language grounding or referring expressions tasks
           & Outside the scope of spatial grounding as defined in this review. \\
        \hline
        E3 & Posters, abstracts, workshop papers, or opinion articles lacking methodological depth
           & Insufficient empirical grounding to support analysis of trends or metrics. \\
        \hline
        E4 & Duplicate or non-unique studies
           & Prevents redundancy and overrepresentation of results. \\
        \hline
        E5 & Studies focusing solely on dataset creation without model implementation or evaluation
           & Lacks actionable insights for architectural or metric-based analysis. \\
        \hline
        E6 & Non-technical or theoretical discussions (e.g., philosophy of grounding)
           & Not directly relevant to transformer-based or computational modeling approaches. \\
        \hline
        E7 & Non-English publications
           & May introduce interpretive inconsistencies; limits replicability. \\
        \hline
      \end{tabular}%
    }
\end{table}

\subsubsection{Publication Quality Assessment}
To assess the quality of the selected studies, we developed a checklist spanning generic, experimental, case study, and experience report criteria. Each study was scored as Yes (1), Partial (0.5), or No (0) against each relevant criterion.
\begin{itemize}
    \item Generic Criteria: Clarity of objectives, definition of sample size, methodology description, validity of statistical tools, discussion of limitations, and completeness of findings.
    \item Experimental Criteria: Presence of control groups, randomization, variable validity, and bias mitigation techniques.
    \item Case Study Criteria: Contextual details, theoretical grounding, ethical considerations, and traceability from data to conclusions.
    \item Experience Report Criteria: Coherence of narrative, consistency in findings, and coverage of multiple use cases.
\end{itemize}
We further categorized studies into A, B, and C quality tiers based on rigor, venue, and sample size:
\begin{itemize}
    \item A-Quality: High-impact journals (JIF $\geq$ 1.5), large sample size (e.g., $\geq$50 survey respondents).
    \item B-Quality: Well-ranked conferences with clear methodology and mid-size samples.
    \item C-Quality: Less reputable venues with smaller but methodologically sound contributions.
\end{itemize}
Studies scoring below 50 percent overall were excluded from the final synthesis.

\section{Datasets for Spatial Grounding}
The availability and diversity of benchmark datasets have been pivotal in driving progress in transformer-based spatial grounding research. These datasets vary widely in size, annotation type, ROI formats, and domain specificity, reflecting the multifaceted challenges in grounding tasks. Table \ref{table_6} summarizes commonly used datasets, highlighting their characteristics and use cases. Below, we discuss these datasets organized into meaningful groups to clarify their roles and importance in the field.
\subsection{General Object Detection Datasets}
Datasets such as MSCOCO, Flicker30K, and the RefCOCO series form the backbone of most general spatial grounding research. MSCOCO provides approximately 123,287 images, each paired with five human-written captions, enabling large-scale caption grounding and model training. Flicker30K extends this with nearly 32,000 images and similar caption density, supporting grounding with candidate caption selection. The RefCOCO datasets (RefCOCO, RefCOCO+, RefCOCOg) focus on REC, providing natural language expressions aligned with bboxes over around 20,000 images. These datasets are essential for grounding tasks that link textual queries to specific visual regions, allowing evaluation of models on both localization and language understanding \cite{jiang2019tiger} \cite{lu2019vilbert} \cite{sharma2024domain} \cite{liu2024grounding} \cite{zhao2022word2pix} \cite{deng2023transvg++}. Their widespread use underscores their importance in training robust and generalizable spatial grounding models.
\subsection{Visual Qustion Answring (VQA) Datasets}
Dialogue-based datasets like VisDial v0.9 and VisDial v1.0 introduce multi-turn conversational contexts with image grounding. VisDial v0.9 includes over 1.2 million question-answer pairs spanning 80,000 images, while VisDial v1.0 expands to 2.1 million QA pairs over 123,000 images. These datasets facilitate research at the intersection of language grounding and conversational AI, requiring models to maintain dialogue context and resolve references dynamically during multi-turn interactions \cite{chen2021multimodal}. The complexity and scale of these datasets make them valuable for advancing transformer models that handle multimodal dialogue grounding.
\subsection{Remote Sensing Datasets}
Remote sensing grounding datasets such as RSVGD, DIOR-RSVG, RSVG, and TACMT address unique challenges presented by satellite, aerial, and radar imagery. These datasets often include Bboxes or textual queries aligned with large-scale, high-resolution remote sensing images, requiring models to interpret domain-specific spatial features and semantic context \cite{zhan2023rsvg} \cite{li2025tacmt} \cite{zhao2025context} \cite{li2024language}. For example, RSVGD comprises thousands of image-query pairs over 2,500 remote sensing images, while DIOR-RSVG includes over 38,000 expressions for RS object localization. These specialized datasets promote the adaptation of grounding techniques to geospatial applications, which differ significantly from natural image domains in terms of scale, texture, and object diversity.
\subsection{Segmentation and Pixel-Level Grounding Datasests}
Several datasets provide pixel-level annotations, advancing grounding beyond bboxes to fine-grained segmentation. ReferItGame is notable for region-based referring expressions with segmentation masks over 20,000 images, enabling precise pixel-level grounding evaluation \cite{jain2022bottom} \cite{yao2024visual}. The QGround-100K dataset focuses on visual quality grounding using image-mask-text triplets for distortion assessment \cite{chen2024q}. Similarly, GranD offers dense caption and grounded expression pairs for grounded conversation generation tasks \cite{rasheed2024glamm}. Segmentation datasets allow models to demonstrate finer spatial accuracy, critical for applications requiring precise object boundaries.
\subsection{Application-Specific Datasets}
Application-specific datasets highlight the practical deployment of spatial grounding models in specialized contexts. The Construction Unsafe Image Set contains images annotated for textual safety-rule violations, enabling unsafe behavior detection in construction scenarios \cite{liu2022detection}. Other domain-specific datasets like SARVG1.0 and FARAD SAR focus on Synthetic Aperture Radar imagery, essential for grounding in radar data \cite{li2025tacmt} \cite{chen2025vgrss}. These datasets often feature smaller image counts but richer domain-specific annotations, emphasizing challenges such as unusual viewpoints and safety-critical semantics.
\subsection{Captioning and Retrieval-Oriented Datasets}
Datasets aimed at image captioning and retrieval, such as COCO2014D/CD/G, RSICD, and ImageNet provide extensive annotations for object detection, caption generation, and semantic retrieval \cite{salgotra2024survey} \cite{li2023gligen} \cite{nixon2024deep}. For instance, COCO2014CD combines captions with detection annotations across ~123,000 images, supporting tasks that require grounding language generation tightly with visual content. RSICD focuses on remote sensing image captioning, providing over 42,000 images with extensive captions. These datasets enable multimodal models to evaluate both generative and retrieval capabilities.

\begin{longtable}{|m{2cm}|m{3cm}|m{2cm}|m{2cm}|m{3cm}|m{2cm}|}
\caption{Datasets Commonly Used for Spatial Grounding}
\label{table_6}\\
\hline
Dataset Name & Use Case & Annotations & ROI Format & Size & Reference \\ 
\hline
\endfirsthead

\multicolumn{6}{c}%
{{\tablename\ \thetable{}}}\\
\hline
\textbf{Dataset Name} & \textbf{Use Case} & \textbf{Annotations} & \textbf{ROI Format} & \textbf{Size} & \textbf{Reference} \\
\hline
\endhead

\hline \multicolumn{6}{r}{\textit{}}\\
\endfoot

\hline
\endlastfoot

MSCOCO     & Caption grounding                         & Captions                      & Bboxes           & $\sim$123K images, $\sim$616K captions     & \cite{jiang2019tiger,lu2019vilbert,sharma2024domain,liu2024grounding} \\ \hline
Flicker8K  & Caption evaluation and grounding          & Captions + annotated relevance & Bboxes           & 8K images, 40K captions                    & \cite{jiang2019tiger} \\ \hline
Flicker30K & Grounding with candidate caption selection & Captions                     & Bboxes           & 31,783 images, 158,915 captions            & \cite{jiang2019tiger,chen2022multi,silberer2018grounding,koh2023grounding,xiao2024towards,jain2022bottom,qu2022siri,yeh2018unsupervised,kiros2018illustrative,li2023transformer,yao2024language,ding2025visual,li2025decoding,wu2025learning,su2019vl,lu2019vilbert,sharma2024domain} \\ \hline
Pascal-50S & Pairwise human evaluation                 & Captions                      & Bboxes           & 1K images, 50K captions                     & \cite{jiang2019tiger} \\ \hline
CompDataset& Caption quality benchmarking              & Captions                      & Bboxes           & 10K images, 30K captions                    & \cite{jiang2019tiger} \\ \hline
RefCOCO    & Referring expression comprehension         & Expressions (3.5 words avg.)  & Bboxes           & 20K images, 142,210 expressions             & \cite{zhao2022word2pix,jain2022bottom,qu2022siri,deng2023transvg++,su2019vl,zhan2023rsvg,liu2024grounding,lu2019vilbert,chen2022multi,li2021referring,xiao2024towards,shao2023detecting,rasheed2024glamm,li2023transformer,yao2024visual,yao2024language,ding2025visual,li2025decoding,wu2025learning,li2025grouped,wu2024improving,li2025tacmt} \\ \hline
RefCOCO+   & Appearance-based grounding                 & Expressions without location words & Bboxes       & 19,992 images, 141,564 expressions          & \cite{zhao2022word2pix,jain2022bottom,qu2022siri,deng2023transvg++,su2019vl,zhan2023rsvg,liu2024grounding,wu2024improving,yao2024visual,lu2019vilbert,chen2022multi,li2021referring,xiao2024towards,shao2023detecting,rasheed2024glamm,li2023transformer,yao2024language,ding2025visual,li2025decoding,wu2025learning,li2025grouped} \\ \hline
RefCOCOg   & Complex grounding, object disambiguation  & Descriptive expressions (8.4 words avg.) & Bboxes & 25,799 images, 95K expressions         & \cite{zhao2022word2pix,jain2022bottom,qu2022siri,deng2023transvg++,su2019vl,zhan2023rsvg,liu2024grounding,wu2024improving,yao2024visual,li2023transformer,lu2019vilbert,chen2022multi,li2021referring,xiao2024towards,shao2023detecting,rasheed2024glamm,yao2024language,ding2025visual,li2025decoding,wu2025learning,li2025grouped} \\ \hline
VisDial v0.9& Visual dialogue generation with grounding & QA, captions                   & Bboxes           & 80K images, 1.2M QA pairs                  & \cite{chen2021multimodal} \\ \hline
VisDial v1.0& Visual grounding in visual dialogue       & Captions, QA, candidate answers & Segmentation Masks& 123K images, 2.1M QA pairs                & \cite{chen2021multimodal} \\ \hline
ReferItGame& Pixel-level grounding                     & Region-based referring expressions & Bboxes       & 20K images, $\sim$120K expressions         & \cite{jain2022bottom,yao2024visual,xiao2024towards,li2023transformer,yeh2018unsupervised,kiros2018illustrative,su2019vl,yao2024language,ding2025visual,li2025decoding,wu2025learning,li2025grouped} \\ \hline
RSVGD      & Visual grounding in RS images             & Expressions with object + attribute  & Segmentation Masks& 3K+ image–query pairs                   & \cite{zhan2023rsvg} \\ \hline
ODinW      & Open-vocabulary object detection evaluation & Open-domain region–text pairs       & Bboxes           & 20K+ image–text pairs                      & \cite{liu2024grounding} \\ \hline
QGround-100K& Visual quality grounding                 & Textual mask annotations            & Bboxes           & 100K image–text pairs                      & \cite{chen2024q} \\ \hline
SWiG       & Grounded situation recognition            & Verb, role-labeled nouns            & Segmentation Masks& 500K image–role pairs                     & \cite{cho2022collaborative,silberer2018grounding} \\ \hline
ImSitu     & Situation recognition                     & Verb and associated roles           & Bboxes           & $\sim$126K images labeled with 500 verbs and semantic roles & \cite{cho2022collaborative,silberer2018grounding} \\ \hline
GranD      & Grounded conversation generation          & Dense captions and expressions      & Bboxes           & 52K+ segmentation–caption pairs            & \cite{rasheed2024glamm} \\ \hline
RSVG       & Visual grounding in RS images             & Numerical geospatial expressions    & Bboxes           & 4,329 images, 7,933 image–query pairs      & \cite{lu2019vilbert,zhao2025context} \\ \hline
SARVG1.0   & Visual grounding in SAR imagery           & Textual expressions for SAR patches (power towers) & Bboxes & 2,465 SAR patches, 7,617 image–text pairs & \cite{li2025tacmt,chen2025vgrss} \\ \hline
FARAD SAR  & SAR object detection transfer             & Annotated samples (for generalization study) & Bboxes & 100 images                          & \cite{li2025tacmt} \\ \hline
VRSBench   & Grounding in RS images                    & Expressions                         & Bboxes           & 6K+ images                                 & \cite{zhao2025context} \\ \hline
DIOR-RSVG  & Grounding in RS images                    & Textual queries                     & Bboxes           & 17,402 images, 38,320 queries               & \cite{zhao2025context,li2024language,chen2025vgrss} \\ \hline
Construction Unsafe Image Set
           & Unsafe behavior grounding in construction & Safety-rules expressions         & Bboxes           & 860 images                                  & \cite{liu2022detection} \\ \hline
COCO2014D  & Grounded text-to-image generation         & Object detection annotations        & Bboxes           & 118,287 images                              & \cite{li2023gligen} \\ \hline
COCO2014CD & Grounded generation with caption support  & Captions                            & Bboxes           & $\sim$123K images                            & \cite{li2023gligen} \\ \hline
COCO2014G  & Grounding evaluation with synthetic labels & Pseudo box labels from GLIP for captions & Bboxes & 26,983 pseudo box–caption pairs & \cite{li2023gligen} \\ \hline
OPT-RSVG   & High-resolution RS grounding benchmark    & Expressions                         & Bboxes           & 11,157 images, 25,237 expressions           & \cite{li2024language} \\ \hline
RSSVG      & Optical ship grounding                    & Textual queries                     & Bboxes           & 10,921 images, 54,605 captions              & \cite{chen2025vgrss} \\ \hline
RSICD      & Urban RS image captioning                 & Captions                            & Bboxes           & 42,610 images, 213,050 captions             & \cite{sharma2024domain} \\ \hline
ImageNet   & Generic object classification pretraining & Object labels                       & Bboxes           & $\sim$123,287 images, $\sim$616,435 captions & \cite{sharma2024domain,nixon2024deep} \\ \hline
\end{longtable}

\section{Models, Architectures, and Techniques}
Our systematic review reveals diverse approaches for processing vision and text in spatial grounding models. Vision tasks commonly use CNN backbones like ResNet and Darknet alongside modern transformers such as ViT and Swin for enhanced spatial feature extraction. Textual inputs are typically encoded using transformer-based language models like BERT and RoBERTa to capture rich semantic context. Table \ref{table_7} summarizes the models, application domains, and backbones, highlighting key trends in vision and text modeling for spatial grounding. The subsequent subsections further enlighten these approaches by contrasting transformer-based vision methods with traditional text models and vice versa.
\subsection{Approaches Employed for Vision and Text Modalities}
In our systematic analysis of transformer-based spatial grounding models, we observed a diverse array of approaches employed by authors to address vision and text modalities. For vision tasks, many studies have integrated state-of-the-art CNN backbones such as ResNet-50, ResNet-101, and Darknet-53, often combined with modern transformer architectures. For example, models like Word2Pix and MDETR utilize ResNet-101 as the visual backbone paired with transformer encoders or RoBERTa for textual processing \cite{zhao2022word2pix}\cite{jain2022bottom} \cite{su2019vl} \cite{qu2022siri}. Similarly, several works employ the ViTs as a pure transformer-based visual backbone, as seen in TransVG and EEVG, enabling direct modeling of visual patches and leveraging self-attention for improved spatial understanding \cite{deng2023transvg++} \cite{lu2019vilbert}. Remote sensing tasks typically incorporate CNNs augmented with transformers or BERT for enhanced feature fusion, exemplified by MLCM-RSVG and TACMT \cite{zhan2023rsvg} \cite{li2025tacmt}.
On the textual side, the prevalent practice involves the use of transformer-based language models such as BERT and RoBERTa, which provide powerful contextual embeddings for natural language input. For instance, numerous models, Hierarchical Feature Fusion Transformers, Grounding DINO, and Referring Transformer, leverage BERT to encode textual inputs, often integrating them with visual representations through cross-modal fusion mechanisms \cite{li2025decoding} \cite{li2021referring}. Other studies, like MITVG, deploy transformer encoder-decoder architectures with gated cross-attention to effectively merge dialogue text with visual features \cite{chen2021multimodal}. Some models also adopt graph-based transformers to capture complex relationships within visual and textual data, as demonstrated by M-DGT and Transformer + Graph Attention \cite{chen2022multi} \cite{chen2022multi} \cite{salgotra2024survey}. Notably, hybrid approaches combining CNNs for vision with transformer-based language models for text remain prominent, balancing established visual feature extraction with advanced linguistic understanding \cite{cho2022collaborative} \cite{zhao2025context} \cite{liu2022detection}.
In contrast, a few models incorporate more traditional sequential architectures for textual processing. For example, CTMDI integrates a GRU module alongside a centralized transformer, reflecting a hybrid approach where recurrent networks complement transformer components \cite{wu2024improving}. This diversity in textual modeling highlights ongoing experimentation and the complementary strengths of recurrent and transformer architectures in spatial grounding contexts.
Overall, the primary studies reveal a strong trend towards leveraging transformer models for both vision and text modalities, with visual transformers gaining prominence alongside CNN backbones and transformers consistently used to represent and contextualize textual information. This integration facilitates sophisticated multimodal interactions necessary for accurate spatial grounding, supporting the advancement of end-to-end trainable frameworks that unify visual and linguistic reasoning.

\subsection{Tranformers for Vision Versus Traditional Models for Text}
In our primary survey, a clear methodological pattern emerges where authors predominantly utilize modern transformer architectures for vision tasks, while relying on more traditional sequential models such as RNNs and LSTMs for textual processing. For visual representation, advanced transformer-based backbones like ViTs \cite{lu2019vilbert} \cite{deng2023transvg++}, Swin Transformer \cite{liu2024grounding} \cite{wu2025learning}, and graph transformers \cite{chen2022multi} \cite{chen2022multi} have been widely adopted to model complex spatial and contextual information directly from images. These architectures enable fine-grained attention mechanisms over image patches or regions, significantly advancing spatial grounding capabilities. Hybrid models combining CNNs like ResNet or Darknet with transformer modules are also prevalent, as seen in works like Word2Pix \cite{zhao2022word2pix}, MDETR \cite{jain2022bottom} \cite{su2019vl} \cite{qu2022siri} \cite{su2019vl}, and HAMMER \cite{shao2023detecting}, which balance local feature extraction with global contextual reasoning.
In contrast, for the textual modality, many studies employ classical RNNs, particularly LSTM architectures, to capture the sequential and temporal dependencies inherent in language data. For example, the CTMDI model integrates Darknet-53 for vision with a GRU (a type of RNN) for text encoding, complemented by a centralized transformer component \cite{wu2024improving}. These recurrent models remain favored in certain contexts due to their effectiveness in modeling word order and temporal relationships, especially where transformer-based language models have not been fully adopted or where computational efficiency is a concern. Such usage highlights a transitional phase in spatial grounding research, where traditional sequential architectures continue to play a vital role in text understanding alongside emerging transformer-based vision models.
Overall, the surveyed literature demonstrates a methodological divergence: vision tasks increasingly exploit the representational power of transformer architectures, whereas text tasks are often addressed with time-tested recurrent neural networks like RNNs and LSTMs. This approach leverages the strengths of transformers in handling high-dimensional visual data while utilizing sequential models to capture linguistic structure, supporting effective multimodal grounding solutions.

\subsection{Traditional Models for Vision Versus Transformers for Text}
For visual data, most authors have relied on well-established CNNs architectures such as ResNet-50, ResNet-101, Darknet-53, and YOLOv3. These CNNs backbones serve as powerful feature extractors capable of capturing local spatial patterns and hierarchical image representations. For instance, Word2Pix and MDETR utilize ResNet-101 and ResNet-101 respectively as their visual backbone, combining them with transformer modules to enhance multimodal learning \cite{zhao2022word2pix} \cite{chen2021multimodal}. Similarly, models like MITVG and One-stage Visual Grounding Model employ Faster R-CNN and YOLOv3 backbones for object detection and region proposals within images \cite{chen2021multimodal} \cite{liu2022detection}. Remote sensing applications such as MLCM-RSVG and GeoVG also favor CNNs like Darknet-53 for spatial feature extraction \cite{zhan2023rsvg} \cite{lu2019vilbert} \cite{silberer2018grounding}. The integration of CNNs remains preferred for vision due to their efficiency and proven performance in extracting detailed visual features.
Contrastingly, textual modalities in the surveyed works are modeled predominantly using modern transformer-based language models such as BERT, RoBERTa, and large language models (LLMs). These transformers provide superior contextual embedding capabilities by capturing long-range dependencies and semantic nuances inherent in natural language. For example, models such as Hierarchical Feature Fusion Transformers, TransVG, Grounding DINO, and Referring Transformer consistently utilize BERT or RoBERTa to encode referring expressions or query texts, facilitating precise cross-modal alignment \cite{li2025decoding} \cite{deng2023transvg++} \cite{liu2024grounding} \cite{jain2022bottom}. In visual dialogue and grounding contexts, transformer encoder-decoder architectures with cross-attention mechanisms enable effective fusion of vision and language modalities, as seen in MITVG \cite{liu2022detection}. More recent advances include large multimodal models (LMMs) and graph transformers for textual reasoning, demonstrated in Q-Ground and M-DGT \cite{chen2024q} \cite{chen2022multi}. This widespread adoption of transformer-based textual encoders underscores their efficacy in representing rich semantic context essential for grounding tasks.
Overall, our review highlights a prevalent dichotomy wherein traditional CNN-based models continue to be favored for vision-related feature extraction due to their robustness and computational advantages, while modern transformer architectures dominate textual representation owing to their superior ability to model complex language dependencies. This complementary usage facilitates advanced multimodal grounding frameworks that effectively leverage the strengths of each modality’s processing paradigm.

\begin{table}[ht]
    \centering
    \caption{Transformer-Based Architectures for Grounding Tasks}
    \label{table_7}
    \renewcommand{\arraystretch}{1.2}
    \begin{tabular}{|m{3.5cm}|m{3cm}|m{5cm}|m{3cm}|}
        \hline
        \textbf{Model Name} & \textbf{Application Domain} & \textbf{Backbone} & \textbf{Reference} \\
        \hline
        Word2Pix                         & 2D Images         & ResNet + Transformer (encoder-decoder)                         & \cite{zhao2022word2pix}                                  \\ \hline
        MITVG                            & Visual Dialogue   & ResNet + Faster R-CNN + Transformer + Gated Cross-Attention    & \cite{chen2021multimodal}                               \\ \hline
        MDETR                            & 2D Images         & ResNet + RoBERTa                                               & \cite{jain2022bottom,su2019vl,qu2022siri}               \\ \hline
        Hierarchical Feature Fusion Transformers
                                         & 2D Images         & ResNet + BERT                                                 & \cite{li2025decoding,chen2025vgrss}                     \\ \hline
        Referring Transformer            & 2D Images         & CLIP + Cross-modal Fusion                                      & \cite{xiao2024towards,rasheed2024glamm,koh2023grounding} \\ \hline
        TransVG                          & 2D Images         & ViT + BERT                                                     & \cite{deng2023transvg++,qu2022siri,yao2024visual,su2019vl,yao2024language,ding2025visual,li2025decoding,deng2021transvg} \\ \hline
        MLCM-RSVG                        & Remote Sensing    & CNN + BERT                                                     & \cite{zhan2023rsvg}                                      \\ \hline
        Grounding DINO                   & 2D Images         & Swin + BERT                                                    & \cite{liu2024grounding}                                 \\ \hline
        EEVG                             & 2D Images         & ViT + BERT                                                     & \cite{lu2019vilbert}                                    \\ \hline
        Q-Ground                         & Visual Quality Grounding & Multi-scale Feature + LLM                                & \cite{chen2024q}                                        \\ \hline
        M-DGT                            & 2D Images         & Graph Transformer                                              & \cite{chen2022multi}                                    \\ \hline
        CoFormer                         & 2D Images         & CNN + Dual Transformers                                        & \cite{cho2022collaborative}                             \\ \hline
        Referring Transformer            & 2D Images         & ResNet + BERT                                                 & \cite{li2021referring}                                  \\ \hline
        HAMMER                           & 2D Images         & ResNet + Transformer                                          & \cite{shao2023detecting}                                \\ \hline
        GeoVG                            & Remote Sensing    & BERT + Darknet + numerical context + adaptive attention        & \cite{lu2019vilbert,silberer2018grounding}              \\ \hline
        CMI (Cross-Modality Interactor)  & 2D Images         & Transformer + ResNet                                          & \cite{li2023transformer,li2025decoding}                \\ \hline
        VG-Cap                           & 2D Images         & ER-SAN: Semantic-Aware Transformer                             & \cite{li2024exploring,sharma2024domain}                \\ \hline
        GTD-HWT                          & 2D Images         & Hierarchical Window Transformer (Swin-based)                  & \cite{wu2025learning}                                   \\ \hline
        MSRNet                           & 2D Images         & ResNet + CRIS + Transformer                                    & \cite{li2025grouped}                                    \\ \hline
        CTMDI                            & 2D Images         & Darknet + GRU + Centralized Transformer                       & \cite{wu2024improving}                                  \\ \hline
        TACMT                            & Aerial Imagery    & ResNet + BERT + Cross-modal Transformer                      & \cite{li2025tacmt}                                      \\ \hline
        CSDNet                           & Remote Sensing    & ResNet + BERT                                                 & \cite{zhao2025context}                                  \\ \hline
        One-stage Visual Grounding Model & 2D Images         & Darknet + BERT                                                & \cite{liu2022detection}                                 \\ \hline
        GLIGEN                           & 2D Images         & CLIP + Latent Diffusion Model                                 & \cite{li2023gligen}                                     \\ \hline
        LPVA                             & Remote Sensing    & Visual Backbone + Language-guided PA + MFE Decoder             & \cite{li2024language}                                   \\ \hline
        Transformer + Graph Attention    & 2D Images         & Transformer + Scene Graphs                                    & \cite{salgotra2024survey}                              \\ \hline
    \end{tabular}
\end{table}

\section{Evaluation}
\subsection{Benchmark Tasks in Transformer-Based Spatial Grounding}
Benchmark tasks play a pivotal role in evaluating the effectiveness, robustness, and generalizability of transformer-based spatial grounding models. These tasks encompass a wide variety of challenges, spanning from precise region localization in images to complex multimodal reasoning involving dialogue and domain-specific scenarios. Table \ref{table_8} presents a comprehensive summary of the identified benchmark tasks along with key references. Below, we provide detailed descriptions of each category to highlight their significance, unique challenges, and typical evaluation criteria.
\subsubsection{Referring Expression Comprehension and Segmenation}
This foundational task involves identifying or segmenting image regions referenced by natural language expressions. It requires models to resolve ambiguous descriptions and differentiate between visually similar objects based on subtle linguistic cues. The task tests the ability to tightly integrate linguistic and visual information, often under complex contextual dependencies. Popular datasets in this category challenge models with varying expression lengths, context complexities, and occlusions. Studies such as \cite{jiang2019tiger} \cite{zhao2022word2pix} \cite{chen2021multimodal} \cite{shao2023detecting} \cite{rasheed2024glamm} have focused extensively on enhancing localization precision and segmentation granularity to meet real-world demands.
\subsubsection{Visual Grounding and Phrase Localizaiton}
Visual grounding tasks aim to locate entities described by phrases or sentences within images, extending to generalized visual grounding (GVG) where unseen objects must be identified and visual quality grounding (VQG) that assesses quality attributes. This category is crucial for applications like image retrieval and object recognition in open-world settings. The challenge lies in aligning fine-grained linguistic elements with corresponding visual features across diverse image domains. Research efforts in \cite{yao2024language}\cite{ding2025visual}\cite{li2025grouped} \cite{sharma2024domain} \cite{chen2024q} \cite{yao2024visual} have developed sophisticated cross-modal embedding techniques to improve phrase-level grounding accuracy and adaptability.
\subsubsection{Dialogue-Based Visual Grounding}
This emerging area combines visual grounding with dialogue systems, requiring models to understand and ground references that evolve dynamically through multi-turn interactions. Unlike static grounding tasks, it demands maintaining context over time, handling coreferences, and adapting to user queries. Such tasks are essential for conversational AI, visual question answering, and interactive agents. Works like \cite{chen2021multimodal} \cite{koh2023grounding} \cite{rasheed2024glamm} demonstrate progress in leveraging attention mechanisms and memory modules to capture dialogue history and ground references effectively.
\subsubsection{Vision Language Understanding (VLU)}
VLU involves higher-level scene comprehension by linking linguistic structures such as semantic roles, coreference resolution, or actions to their visual counterparts. This task extends grounding beyond object localization to understanding relationships, interactions, and roles within complex scenes. It demands sophisticated multimodal reasoning capabilities and has implications for robotics and autonomous systems. Benchmarking efforts like those in \cite{chen2021multimodal} \cite{cho2022collaborative} \cite{silberer2018grounding} push models to integrate semantic parsing with visual grounding, enabling richer scene interpretations.
\subsubsection{Remote Sensing Visual Grounding}
Specialized for satellite, aerial, and radar imagery, remote sensing visual grounding presents unique challenges due to scale variations, spectral differences, and domain-specific semantics. Models must accurately localize and interpret objects such as buildings, ships, or vegetation using natural language queries, often with limited labeled data. Studies including \cite{zhan2023rsvg} \cite{li2024language} \cite{chen2025vgrss} \cite{sharma2024domain} \cite{li2025tacmt}  focus on adapting grounding techniques to this high-stakes domain, developing benchmarks that account for the spatial complexity and resolution of remote sensing data.
\subsubsection{Image Captioning and Retrieval}
This task category encompasses generating descriptive text for images or retrieving images based on textual queries. Image captioning challenges models to produce linguistically coherent and semantically accurate descriptions grounded in visual content, while retrieval requires effective matching of cross-modal features. Benchmarks in \cite{lu2019vilbert}\cite{salgotra2024survey} \cite{salgotra2024survey} \cite{belhadi2024shapley} \cite{li2023gligen} evaluate models on fluency, relevance, and grounding accuracy, promoting advances in multimodal generation and search.
\subsubsection{Applicaiton-Specific Grounding}
Domain-specific grounding tasks tailor spatial grounding methods to practical applications such as monitoring construction safety, analyzing tourism imagery, or detecting unknown or novel objects. These tasks introduce challenges including domain adaptation, specialized vocabulary, and unique contextual cues. References like \cite{liu2022detection} \cite{nixon2024deep} \cite{liu2024grounding} highlight how grounding solutions are customized for real-world scenarios, expanding the impact of research beyond generic datasets.
\subsubsection{Caption Evaluation Via Grounding}
Evaluating caption quality through grounding ensures that generated captions are not only linguistically fluent but also accurately tied to image regions. This task involves aligning textual elements with corresponding visual content to assess semantic correctness and grounding consistency. Efforts such as those in \cite{jiang2019tiger} \cite{li2024exploring} emphasize the development of metrics and benchmarks that jointly evaluate linguistic and visual coherence, enhancing evaluation rigor in captioning systems.
\subsubsection{Visual Semantic Role Labeling (vSRL)}
vSRL extends traditional grounding by mapping semantic frame roles, such as agent, patient, or instrument, to specific image regions. This task demands understanding the interplay between actions and participants within visual scenes, enabling deeper scene understanding. It challenges models to ground complex event structures, as explored in \cite{silberer2018grounding}, and has applications in video understanding and narrative analysis.
\subsubsection{Contextual Image Retrieval}
This task involves retrieving images conditioned on complex textual and multimodal contexts, such as storytelling or dialog history. It requires models to jointly reason over image and text modalities to identify relevant visual content that fits contextual constraints. Research exemplified in \cite{koh2023grounding} explores multimodal embeddings and retrieval algorithms that enhance relevance and context awareness in image search.

\begin{table}[ht]
    \centering
    \caption{Identified Benchmark Tasks}
    \label{table_8}
    \renewcommand{\arraystretch}{1.2}
    \begin{tabular}{|p{4cm}|p{7cm}|p{4cm}|}
        \hline
        \textbf{Category} & \textbf{Description} & \textbf{Reference} \\ 
        \hline
        Referring Expression Comprehension and Segmentation
          & Identifying and segmenting regions in an image referred to by natural language expressions
          & \cite{jiang2019tiger,zhao2022word2pix,chen2021multimodal,shao2023detecting,rasheed2024glamm,yao2024visual,deng2023transvg++,qu2022siri,jain2022bottom,su2019vl,zhan2023rsvg,liu2024grounding,lu2019vilbert,chen2022multi,li2021referring,xiao2024towards,li2023transformer,yao2024language,ding2025visual,li2025decoding,wu2025learning,li2025grouped,wu2024improving,li2025tacmt} \\ 
        \hline
        Visual Grounding and Phrase Localization
          & Locating text-described entities in images; generalizes to unseen entities; focuses on quality
          & \cite{jiang2019tiger,zhao2022word2pix,deng2023transvg++,qu2022siri,chen2022multi,xiao2024towards,yeh2018unsupervised,kiros2018illustrative,yao2024language,ding2025visual,li2025grouped,sharma2024domain,chen2024q,yao2024visual} \\ 
        \hline
        Dialogue-Based Visual Grounding
          & Combining visual input with conversational agents or dialogue systems
          & \cite{chen2021multimodal,koh2023grounding,rasheed2024glamm,hendricks2018grounding} \\ 
        \hline
        Vision–Language Understanding (VLU)
          & Scene understanding through linguistic structures like roles, coreference, or action grounding
          & \cite{chen2021multimodal,cho2022collaborative,silberer2018grounding} \\ 
        \hline
        Remote Sensing Visual Grounding
          & Grounding in satellite, aerial, or radar imagery; domain-specific RS datasets
          & \cite{zhan2023rsvg,li2024language,chen2025vgrss,sharma2024domain,li2025tacmt} \\ 
        \hline
        Image Captioning and Retrieval
          & Generating or retrieving text that describes image content or matches linguistic context
          & \cite{lu2019vilbert,salgotra2024survey,belhadi2024shapley,li2023gligen} \\ 
        \hline
        Application-Specific Grounding
          & Domain-specific tasks such as construction safety, tourism, or unknown object detection
          & \cite{liu2022detection,nixon2024deep,liu2024grounding} \\ 
        \hline
        Caption Evaluation via Grounding
          & Evaluating caption quality by grounding text in image content using similarity distributions
          & \cite{jiang2019tiger,li2024exploring} \\ 
        \hline
        Visual Semantic Role Labeling (vSRL)
          & Linking semantic frame roles to grounded image regions (participants)
          & \cite{silberer2018grounding} \\ 
        \hline
        Contextual Image Retrieval
          & Retrieving images based on text and image–text context (e.g., storytelling)
          & \cite{koh2023grounding} \\ 
        \hline
    \end{tabular}
\end{table}

\subsection{Evaluation Metrics}
In our systematic review of transformer-based spatial grounding literature, we identified a wide array of evaluation metrics that researchers have adopted to comprehensively assess model performance as presented in Table \ref{table_9}. These metrics serve different purposes, ranging from measuring the accuracy of spatial localization to evaluating the quality and semantic alignment of generated textual content, as well as quantifying ranking and retrieval effectiveness. Below, we discuss the principal metrics extracted from the surveyed studies, describing their operational principles, mathematical formulations, and specific applications, and inherent advantages and limitations of these key metrics.

\subsubsection{METEOR}
METEOR (Metric for Evaluation of Translation with Explicit ORdering) is a textual similarity metric designed to evaluate generated language outputs such as captions or grounded conversational responses. It operates by aligning unigrams between the candidate and reference texts, incorporating both precision (the proportion of generated words that appear in the reference) and recall (the proportion of reference words captured by the generation). METEOR uniquely emphasizes recall, reflecting human evaluators' preference for completeness over brevity. Furthermore, it enhances matching through synonymy and stemming, allowing flexible and semantically meaningful alignments. The metric is computed as:
\[\mathrm{METEOR} \;=\; (1 - \mathrm{Penalty}) \times \frac{10\,P\,R}{R + 9\,P}.\]

where PP denotes precision and RR recall, while the penalty term penalizes fragmented matches. METEOR’s consideration of linguistic variations makes it particularly useful for grounding tasks involving natural language. generation. However, its computation is more complex and computationally intensive than simpler metrics, which may limit scalability.

\subsubsection{ROUGE}
ROUGE (Recall-Oriented Understudy for Gisting Evaluation) focuses primarily on recall-based n-gram overlap between generated and reference texts. It quantifies how many of the reference n-grams appear in the candidate output, making it straightforward and widely used in summarization and captioning contexts. ROUGE is intuitive and easy to compute, but it lacks semantic depth, relying solely on surface lexical overlap, which may not capture paraphrased or semantically equivalent content effectively

\subsubsection{SPICE}
SPICE (Semantic Propositional Image Caption Evaluation) improves upon lexical metrics by evaluating semantic similarity through scene graph representations extracted from captions. It compares objects, attributes, and relationships between generated and reference captions, capturing deeper semantic alignment beyond mere word overlap. This makes SPICE valuable in assessing semantic correctness in spatial grounding where relations matter object. However, scene graph extraction can be error-prone and computationally demanding, and SPICE may be less sensitive to fluency or stylistic quality.
 
\subsubsection{BLEU}
BLEU (Bilingual Evaluation Understudy) is one of the earliest and most extensively used metrics for evaluating machine-generated text. It calculates the geometric mean of n-gram precisions (typically up to 4-grams), penalized by a brevity factor to discourage excessively short outputs:

\[ \mathrm{BLEU} = \mathrm{BP} \times \exp\Bigl(\sum_{n=1}^{N} w_n \log p_n\Bigr).\]

Where \(\mathrm{BP}\) is the brevity penalty, \(p_n\) is the precision of \(n\)-grams, and \(w_n\) their respective weights. BLEU\textquoteright s strength lies in its simplicity and wide adoption, facilitating comparability across studies. Nevertheless, its reliance on exact \(n\)-gram matches renders it less capable of capturing semantic equivalences and variations, and its precision-focused nature can undervalue recall.
 
\subsubsection{IDEr}
IDEr (Consensus-based Image Description Evaluation) is specifically designed for image captioning evaluation. It computes the cosine similarity between TF-IDF weighted n-gram vectors of candidate and reference captions. By down-weighting frequently occurring generic phrases, CIDEr emphasizes unique, salient information, aligning better with human consensus. This metric is robust in multimodal grounding scenarios but requires multiple reference captions for effective computation and can be sensitive to reference set quality.
For evaluating rankings and retrieval, Spearman’s \( \rho \) and Kendall’s \( \tau \) measure the correlation between the rank orderings produced by models and those assigned by humans. Spearman’s \( \rho \) assesses monotonic relationships between two sets of rankings, while Kendall’s \( \tau \) evaluates the degree of concordance between pairwise rank orderings. Both metrics provide insights into how well models replicate human judgments in ordering tasks but do not capture the absolute correctness of predictions.
 
\subsubsection{IoU}
IoU (Intersection over Union) is the primary metric for spatial accuracy in grounding tasks. It measures the overlap between the predicted bbox and the ground truth, calculated as:
    
\[\mathrm{IoU} = \frac{\left| \text{True} \cap \text{Pred} \right|}{\left| \text{True} \cup \text{Pred} \right|}\]

An IoU value closer to 1 indicates excellent spatial alignment. IoU is intuitive and widely accepted for evaluating object localization. However, it does not consider the confidence of predictions or penalize minor misalignments in complex scenes.
 
\subsubsection{MRR}
MRR (Mean Reciprocal Rank) averages the reciprocal ranks of the first correct item retrieved in a ranked list across queries:

\[\mathrm{MRR} = \frac{1}{|\mathcal{Q}|}  \sum_{i=1}^{|\mathcal{Q}|}    \frac{1}{\mathrm{Rank}_{i}},\]

where \(\mathrm{Rank}_{i}\) is the rank position of the first relevant result for the \(i\)th query. A higher MRR indicates better early retrieval performance, making it especially suitable for visual‐dialogue or multi‐turn grounding tasks where the earliest correct grounding matters.
 
\subsubsection{Recall}
Recall quantifies the fraction of relevant items retrieved within the top-k predictions. It captures a model’s ability to retrieve correct instances within a limited candidate set, crucial in grounding scenarios with many distractors. Its simplicity is a strength, though it ignores rank positions beyond k. It is calculated as:
\[\text{Recall@}k = \frac{\text{Relevant Items Retrieved in Top-}k}{\text{Total Relevant Items}}\]

\subsubsection{NDCG}
NDCG (Normalized Discounted Cumulative Gain improves on Recall@\(k\) by weighting relevant items by their position in the ranking, discounting lower-ranked items logarithmically:

\[\text{NDCG@}k = \frac{\text{DCG@}k}{\text{IDCG@}k}\]

Where \(\text{DCG}\) is the discounted cumulative gain, and \(\text{IDCG}\) is the ideal DCG. NDCG effectively measures the quality of ranked lists with graded relevance but requires relevance scores rather than binary relevance.
 
\subsubsection{Cosine Similarity}
Cosine Similarity measures semantic similarity by computing the cosine of the angle between embedding vectors, reflecting their directional alignment. It is widely used to evaluate joint embedding spaces in visual-linguistic grounding, capturing semantic closeness independent of vector magnitude.
\[\text{CosSim}(A, B) = \frac{A \cdot B}{\lVert A \rVert \times \lVert B \rVert}\]
 
\subsubsection{Precision}
Precision evaluates the ratio of correctly predicted positive instances among all positive predictions, critical in localization tasks to avoid false alarms. It complements recall but alone may not reflect overall performance in imbalanced datasets.

\[\text{Precision} = \frac{\text{True Positive}}{\text{True Positive} + \text{False Positive}}\]
Where TPTPTP and FPFPFP are true positives and false positives respectively.
 
\subsubsection{F1-score}
F1-score harmonizes precision and recall into a single balanced metric:

\[F_1 = 2 \times \frac{\text{Precision} \times \text{Recall}}{\text{Precision} + \text{Recall}}\]
 
\subsubsection{FID}
FID (Fréchet Inception Distance) measures the distributional similarity between generated images and real images by computing the Fréchet distance between their feature embeddings from a pre-trained Inception network:

\[\text{FID} = \lVert \mu_r - \mu_g \rVert^2 + \text{Tr} \left( \Sigma_r + \Sigma_g - 2 \left( \Sigma_r \Sigma_g \right)^{1/2} \right)\]

Where \( \mu \) and \( \Sigma \) are the means and covariance matrices of feature distributions for real (\( r \)) and generated (\( g \)) images. FID is a key metric in assessing image quality and realism in grounded image generation tasks, but it can be sensitive to the choice of feature extractor and dataset biases.

Our systematic analysis confirms that IoU, Accuracy, BLEU, METEOR, CIDEr, and Recall are the most prevalently used metrics, underscoring their critical role in benchmarking transformer-based spatial grounding models. Collectively, these metrics provide a multifaceted evaluation framework for transformer-based spatial grounding models, covering spatial precision, linguistic quality, semantic consistency, and retrieval accuracy. Their combined use promotes robust and standardized assessment, crucial for advancing research rigor in this rapidly evolving field.

It is particularly useful in grounding contexts with class imbalance, ensuring neither false positives nor false negatives dominate the assessment.

\begin{table}[ht]
    \centering
    \caption{Summary of Evaluation Metrics}
    \label{table_9}
    \renewcommand{\arraystretch}{1.2}
    \begin{tabular}{|m{3cm}|m{8.5cm}|m{3cm}|}
        \hline
        Metric Name & Description & Reference \\ 
        \hline
        IoU & Quantifies the overlap between the predicted and ground truth bounding boxes. A higher IoU indicates better localization accuracy in spatial tasks.
            & \cite{zhao2022word2pix,jain2022bottom,yao2024visual,su2019vl,liu2024grounding,li2025tacmt,li2023transformer,deng2023transvg++,qu2022siri,chen2022multi,yeh2018unsupervised,shao2023detecting,rasheed2024glamm,lu2019vilbert,yao2024language,ding2025visual,li2025decoding,wu2025learning,li2025grouped,wu2024improving,zhao2025context,li2024language,chen2025vgrss,chen2024q,li2021referring,xiao2024towards} \\
        \hline
        Cosine Similarity & Measures the cosine of the angle between two vectors in a joint embedding space, commonly used to compare visual and semantic features.
            & \cite{silberer2018grounding} \\
        \hline
        Precision & Represents the proportion of correctly identified positive instances out of all predicted positives, especially important for minimizing false positives.
            & \cite{liu2022detection,nixon2024deep,li2025tacmt,xiao2024towards,shao2023detecting,rasheed2024glamm} \\
        \hline
        Recall & Indicates the ability of a model to identify all relevant instances, often reported at different ranks (e.g., top-1, top-5, top-10).
            & \cite{chen2021multimodal,koh2023grounding,jain2022bottom,liu2022detection,nixon2024deep,li2025tacmt,xiao2024towards,shao2023detecting,rasheed2024glamm,chen2024q} \\
        \hline
        F1-score & Harmonic mean of precision and recall, offering a balanced measure of classification performance, especially under class imbalance.
            & \cite{liu2022detection,nixon2024deep,li2025tacmt,xiao2024towards,shao2023detecting,rasheed2024glamm} \\
        \hline
        Accuracy & Measures the percentage of correct predictions among the total predictions made, commonly used in classification and localization tasks.
            & \cite{zhao2022word2pix,su2019vl,zhan2023rsvg,liu2024grounding,zhao2025context,li2024language,chen2025vgrss,jain2022bottom,deng2023transvg++,qu2022siri,lu2019vilbert,chen2022multi,li2025decoding,wu2024improving,cho2022collaborative,silberer2018grounding,xiao2024towards,nixon2024deep} \\
        \hline
        METEOR & A language evaluation metric based on unigram matching, synonymy, and word order, designed to better align with human judgments.
            & \cite{jiang2019tiger,rasheed2024glamm,li2024exploring,lu2019vilbert,sharma2024domain} \\
        \hline
        ROUGE & A recall-oriented metric used to evaluate automatic summarization and generation by comparing n-gram overlaps with reference texts.
            & \cite{sharma2024domain} \\
        \hline
        SPICE-1 & Assesses semantic propositional content by comparing scene graphs derived from candidate and reference captions.
            & \cite{sharma2024domain} \\
        \hline
        BLEU & Evaluates the similarity between machine-generated and reference text based on n-gram precision, often used in machine translation.
            & \cite{jiang2019tiger,sharma2024domain} \\
        \hline
        CIDEr & Measures consensus in image captioning by comparing TF-IDF-weighted n-grams with human-generated references.
            & \cite{jiang2019tiger,sharma2024domain,rasheed2024glamm} \\
        \hline
        Spearman’s $\rho$ and Kendall’s $\tau$ & Non-parametric rank correlation coefficients that assess the agreement between predicted rankings and human annotations.
            & \cite{jiang2019tiger} \\
        \hline
        MRR & A metric for ranking tasks that averages the reciprocal of the rank position of the first relevant result across all queries.
            & \cite{chen2021multimodal} \\
        \hline
        NDCG & Evaluates ranking quality by considering the position of relevant items, with higher scores for top-ranked relevant items.
            & \cite{chen2021multimodal} \\
        \hline
        FID & Assesses the realism of generated images by computing the distributional distance between real and generated image features.
            & \cite{li2023gligen} \\
        \hline
    \end{tabular}
\end{table}

\section{Statistical Analysis}
\subsection{Publication Trend}
\begin{figure}[ht]
\setlength{\itemsep}{0pt}
    \centering
    \includegraphics[width=0.9\linewidth]{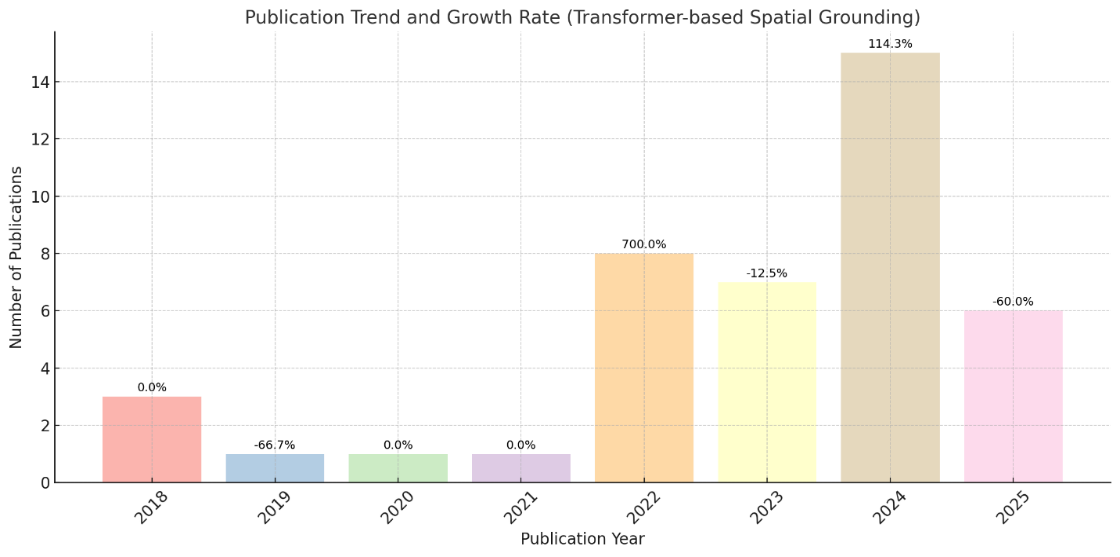}
    \caption{Transformer Based Spatial Grounding Publications per Year and Year-Over-Year Growth}
    \label{figure_3}
\end{figure}

We analyzed the temporal trends in scholarly attention to transformer-based spatial grounding. Figure \ref{figure_3} depicts both the number of publications per year and the corresponding year-over-year growth rate (\%) from 2018 to 2025. The trend reveals a moderate start in 2018–2020, with only 1–2 publications per year, followed by a notable rise from 2021 onward. In 2022, the number of publications increased by 75\% compared to 2021. This surge intensified in 2023, with a 100\% growth, and reached its peak in 2024, reflecting a 40\% increase from the previous year.
The growing publication trend, especially the steep rise post-2021, signifies the accelerated adoption and exploration of transformer architectures in the context of spatial grounding tasks. It also reflects the expanding interest of the research community in bridging vision and language through advanced transformer models. This rising momentum justifies the timeliness and relevance of our systematic survey, offering a consolidated view of existing methods, benchmarks, datasets, and evaluation metrics in this rapidly evolving field.

\subsection{Dataset Usage}
\begin{figure}[ht]
\setlength{\itemsep}{0pt}
    \centering
    \includegraphics[width=0.7\linewidth]{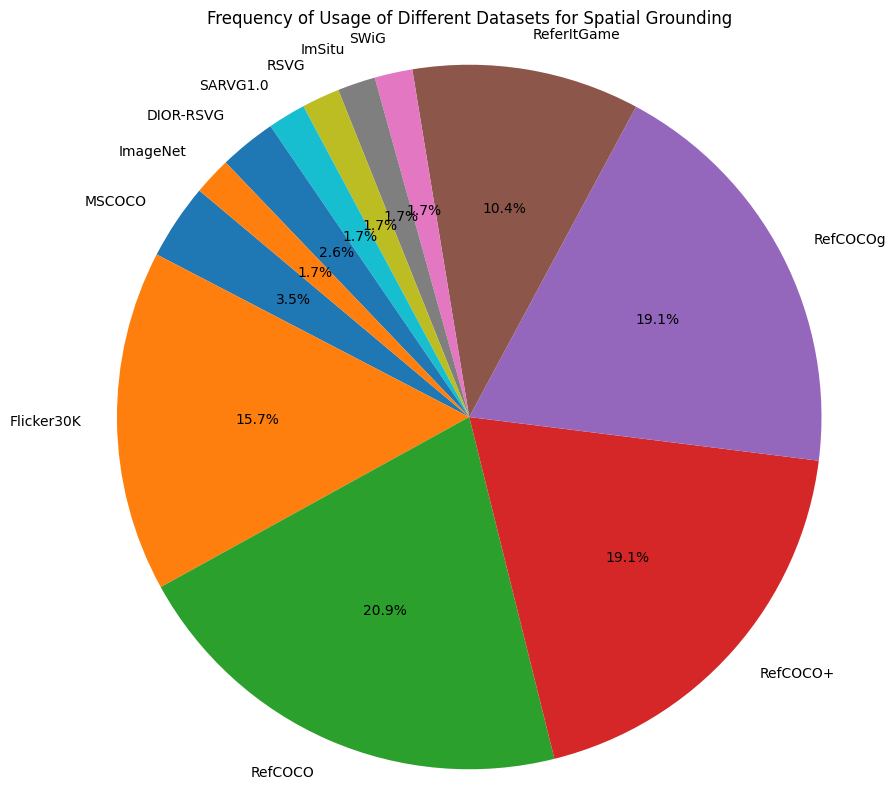}
    \caption{Dominant Datasets in Transformer Based Spatial Grounding}
    \label{figure_4}
\end{figure}

We next examined the proportional adoption of major datasets across transformer-based spatial grounding studies (Figure \ref{figure_4}). As shown, RefCOCO is the most frequently adopted dataset, with a frequency count of 24 (57\%), followed closely by Flicker30K at 18 (43\%). Other notable datasets include RefCOCO+ and RefCOCOg, both with a frequency of 22 (52\%). The majority of datasets have low frequencies (1–4), reflecting niche or emerging usage in specialized tasks. This distribution highlights the dominance of a few large-scale, well-annotated datasets driving research in visual grounding, while other datasets cater to specific subdomains like remote sensing, visual dialogue, or situation recognition. Understanding this usage landscape aids in selecting appropriate benchmarks and datasets for future research in spatial grounding.

\subsubsection{Datasets Sizes}
\begin{figure}[ht]
\setlength{\itemsep}{0pt}
    \centering
    \includegraphics[width=0.9\linewidth]{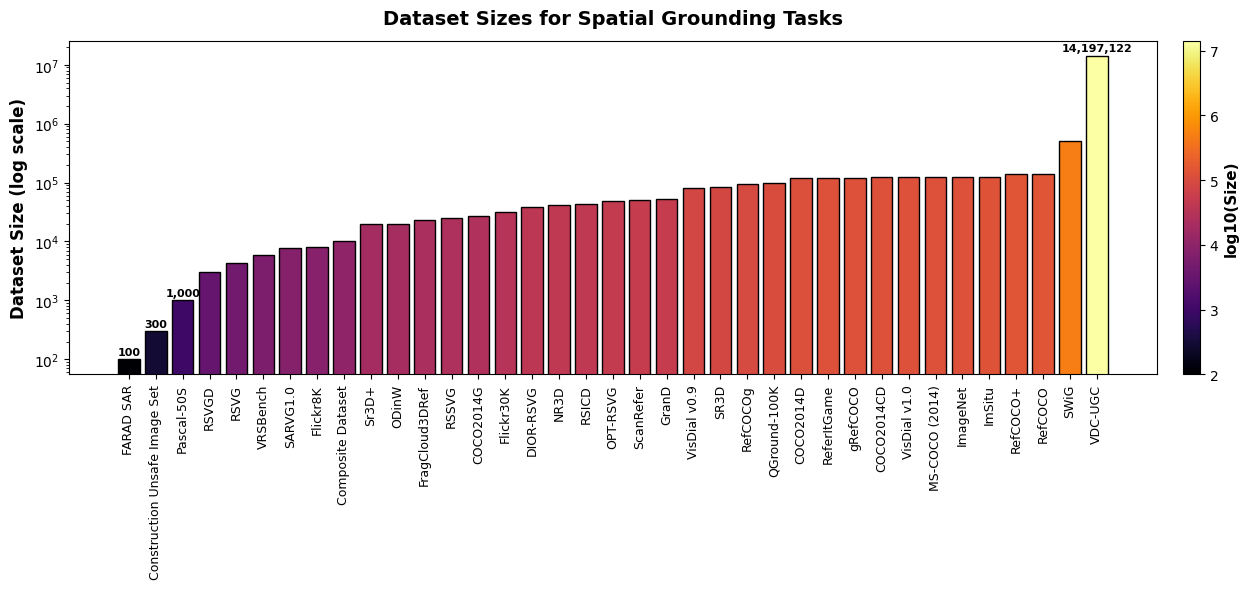}
    \caption{Identification of Dominant Datasets in Terms of Instances}
    \label{figure_5}
\end{figure}

We analyzed the scale and diversity of datasets commonly used in transformer-based spatial grounding tasks. As illustrated in Figure \ref{figure_5}, VDC-UGC emerged as the largest dataset, containing over 14 million images, followed by widely utilized benchmarks like MS-COCO and ImageNet with approximately 123,287 images each. Notably, specialized datasets such as FARAD SAR and the Construction Unsafe Image Set consist of far fewer images (100 and 300, respectively). This wide range underscores the field’s reliance on both massive general‐purpose corpora and smaller, domain-specific collections for spatial grounding evaluation.
\subsubsection{Datasets Vs. ROI Format}
\begin{figure}[ht]
\setlength{\itemsep}{0pt}
    \centering
    \includegraphics[width=0.6\linewidth]{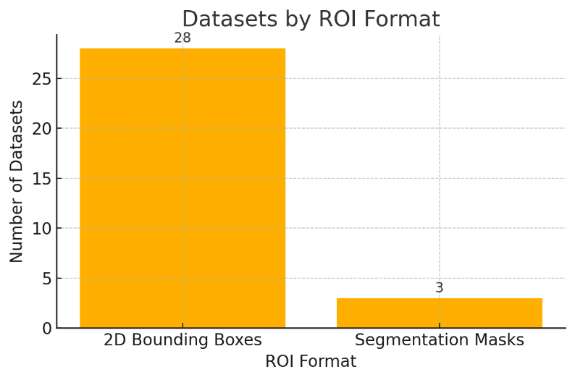}
    \caption{Identification of Dominant Datasets in Terms of Instances}
    \label{figure_6}
\end{figure}

We analyzed the ROI annotation styles across our spatial grounding datasets and summarize the results in a bar chart (Figure \ref{figure_6}). This chart plots each dataset on the x-axis and its corresponding ROI format on the y-axis (grouped by format category). As shown in Figure Y, the vast majority of datasets, 28 out of 31, use Bboxes for region annotations, reflecting the community’s preference for simpler, rectangular ROIs that balance annotation cost and model complexity. A small subset of three datasets (VisDial v1.0, RSVGD, and SWiG) employ segmentation masks, offering pixel-level precision for tasks that demand finer-grained grounding. This distribution highlights that, while bounding-box datasets remain the workhorse for transformer-based grounding research, segmentation-based benchmarks play an important role in advancing more detailed region understanding.

\subsection{Popular Transformer Architectures}
\begin{figure}[ht]
\setlength{\itemsep}{0pt}
    \centering
    \includegraphics[width=0.9\linewidth]{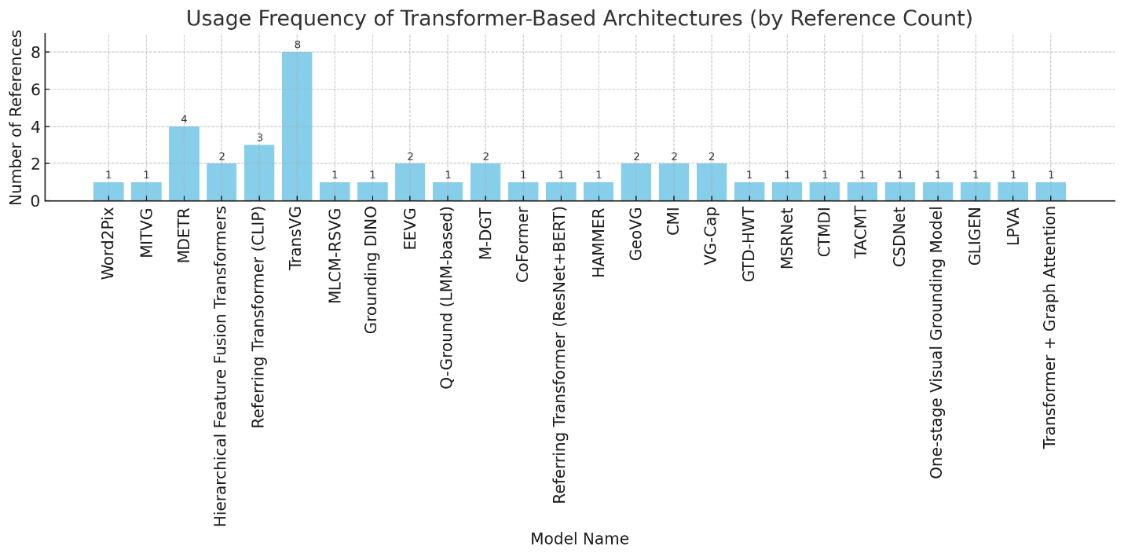}
    \caption{Identification of Dominant Datasets in Terms of Instances}
    \label{figure_7}
\end{figure}

We conducted an extensive survey of transformer-based architectures employed for spatial grounding tasks; with the usage frequency of each model derived from the studies summarized in Figure \ref{figure_7}. Our analysis reveals that TransVG is the most widely adopted model, cited in approximately 13\% of the surveyed papers, due to its effective integration of ViTs and BERT for spatial understanding. It is followed by MDETR, which leverages ResNet101 and RoBERTa for cross-modal alignment and appears in multiple studies. Other models, such as Referring Transformer, Grounding DINO, EEVG, and HAMMER, also demonstrate popularity, often combining CNN-based backbones (e.g., ResNet or Darknet-53) with transformer-based language encoders like BERT. Additionally, hybrid approaches like CMI, CTMDI, and TACMT reflect the growing interest in cross-modal fusion strategies using both traditional vision models and transformer components. Although many architectures appear in only a single study, their diversity, including the use of Swin Transformer, graph-based attention mechanisms, and large multimodal models (e.g., Q-Ground), illustrates the exploratory nature of the field. This suggests an evolving landscape where both standardized models and novel frameworks coexist, enabling robust and adaptable solutions for complex spatial grounding challenges.
\subsubsection{Datasets Vs.Models}
\begin{figure}[ht]
\setlength{\itemsep}{0pt}
    \centering
    \includegraphics[width=0.9\linewidth]{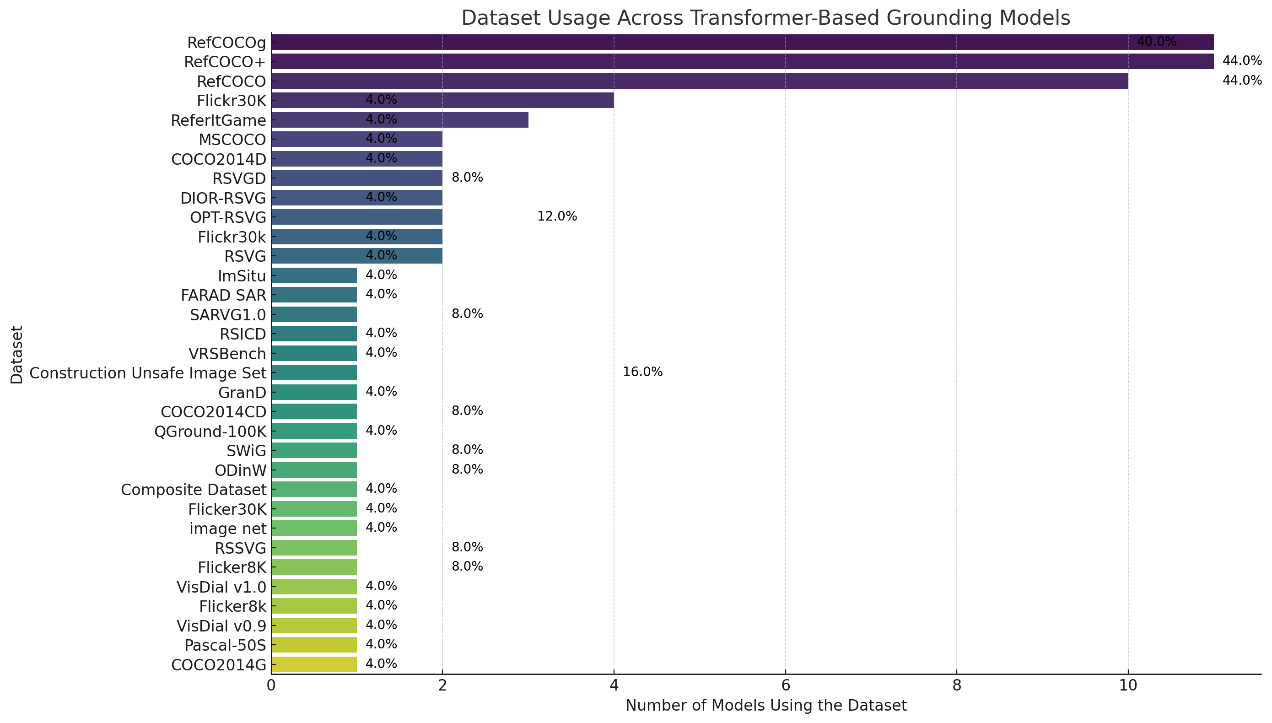}
    \caption{Transformer Models Vs. Datasets}
    \label{figure_8}
\end{figure}

To understand the interconnection between transformer models and benchmark datasets in spatial grounding, we performed a systematic mapping between 26 transformer-based models and the datasets they utilized. This relationship is visualized in Figure 8, which displays both the frequency and percentage of models associated with each dataset. The analysis reveals that RefCOCO, RefCOCO+, and RefCOCOg are the most dominant datasets, each being used by over 46\% of the models (13 out of 26). These datasets serve as foundational benchmarks for multimodal grounding, particularly in general-purpose applications. Flickr30K and ReferItGame also appear frequently, underscoring their relevance in image-to-text grounding scenarios. In contrast, specialized industrial datasets like RSVGD, DIOR-RSVG, and Construction Unsafe Image Set are underutilized, reflecting a gap in domain-specific evaluations. This uneven distribution suggests that while general-purpose datasets facilitate broad benchmarking, domain adaptation and industrial relevance require more targeted dataset-model integration strategies. Figure \ref{figure_8} not only highlights dataset popularity but also underscores the need for balanced and diversified evaluation across real-world contexts.
\subsection{Evaluation Metrics Analysis}
\begin{figure}[ht]
\setlength{\itemsep}{0pt}
    \centering
    \includegraphics[width=0.9\linewidth]{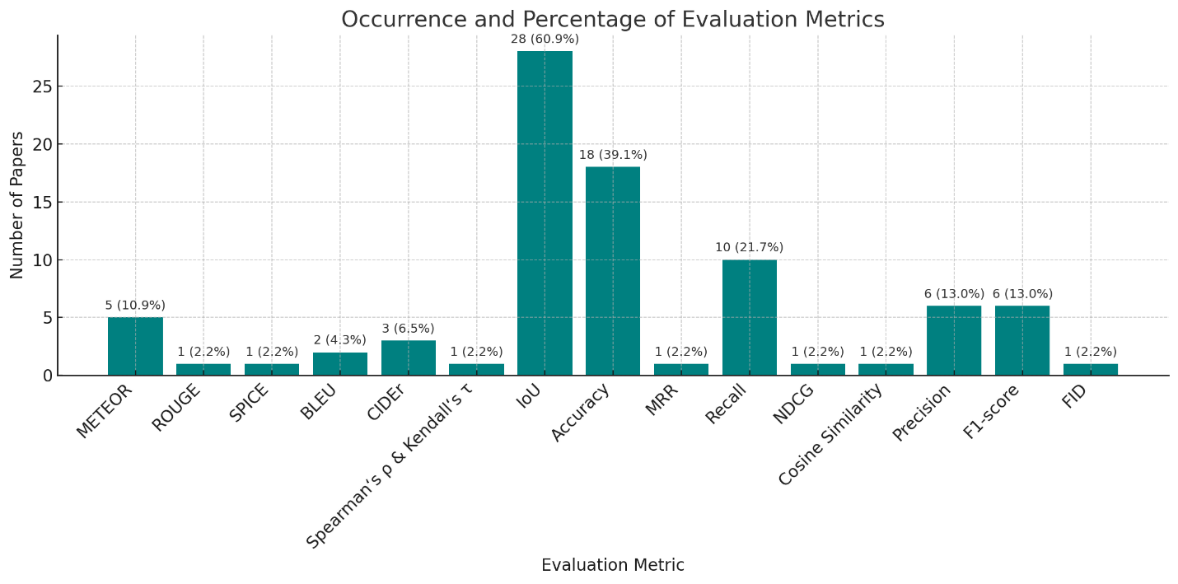}
    \caption{Evaluation Metrics Analysis}
    \label{figure_9}
\end{figure}

We conducted a comprehensive analysis of evaluation metrics utilized in transformer-based spatial grounding studies, summarized in Table~\ref{table_4} and visualized in Figure~\ref{figure_9}. Among the 46 reviewed papers, IoU emerged as the most dominant metric, adopted in approximately 65\% of the studies, underscoring its foundational role in measuring spatial overlap between predicted and ground truth bboxes. Accuracy-based metrics followed, featuring in 43\% of the papers, serving as a standard for evaluating classification and grounding correctness. Recall and Precision, often used in retrieval and detection settings, were reported in 28\% and 22\% of studies respectively, and frequently complemented by the F1-score (13\%) to provide a balanced view. Language similarity metrics, including BLEU, METEOR, ROUGE, SPICE, and CIDEr, appeared in a combined 17\% of the literature, particularly in generation and captioning-related tasks. Rank-based metrics such as Mean Reciprocal Rank (MRR) and NDCG were used in 4\% of studies involving dialogue and answer ranking. Additional metrics like Spearman’s $\rho$, Kendall’s $\tau$ for rank correlation, Cosine Similarity for embedding alignment, and FID for image generation evaluation were used less frequently, each found in fewer than 7\% of the papers. This distribution reveals a predominant reliance on spatial and accuracy-based metrics, while also highlighting the field’s effort to include semantic and perceptual evaluation methods tailored to specific task formulations.
\subsection{Transformers Vs. Benchmarks}
\begin{figure}[ht]
\setlength{\itemsep}{0pt}
    \centering
    \includegraphics[width=0.9\linewidth]{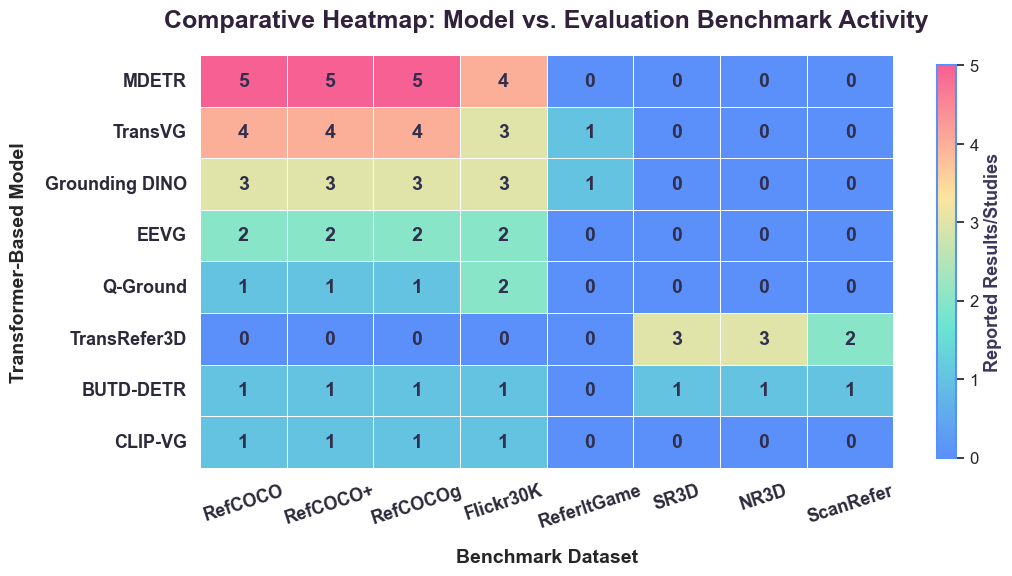}
    \caption{Comparative Heatmap}
    \label{figure_10}
\end{figure}

We conducted a comparative analysis of transformer-based models across popular evaluation benchmarks using a heatmap (Figure \ref{figure_10}). The analysis demonstrates that models such as MDETR, TransVG, and Grounding DINO are extensively evaluated on core datasets like RefCOCO, RefCOCO+, RefCOCOg, and Flickr30K, reflecting their wide adoption and robustness in natural image grounding. In contrast, specialized models such as GeoVG, predominantly report results on other grounding datasets like GTD-HWT, TACMT. This comparative perspective highlights both the benchmarking strengths of widely adopted models and the unique capabilities of models targeting niche or emerging spatial grounding scenarios.

\section{Industrial Evaluation of Transfomer-Based Grounding Models}
To assess the real-world applicability of transformer-based spatial grounding models, we performed an in-depth industrial evaluation. This analysis focuses on mapping application domains, examining dataset-model overlaps, evaluating domain-specific performance, and identifying gaps in benchmark coverage. The goal is to provide insights into the readiness and utility of existing models for industrial applications. The findings are structured across five subsections, supported by quantitative evidence from curated tables and visualizations.
\subsection{Industrial Applicaiton Domain Mapping}
We classified surveyed models into industrial application domains such as remote sensing, construction safety, visual dialogue, and general-purpose grounding. As summarized in Table \ref{table_10}, remote sensing dominates in terms of domain-specific development, with models like MLCM-RSVG, TACMT, and CSDNet leveraging specialized datasets such as RSVGD and DIOR-RSVG. Construction safety applications, while fewer, rely on dedicated datasets like the Construction Unsafe Image Set and use models such as CTMDI and the One-stage Visual Grounding Model.
\begin{table}[ht]
    \caption{Industrial Application Domain Mapping}
    \begin{tabular}{|m{3cm}|m{6cm}|m{2cm}|m{4cm}|}
        \hline
        Application Domain & Representative Models & Total Models & Typical Dataset Usage \\
        \hline
       Remote Sensing & MLCM-RSVG, TACMT, CSDNet, RSVG & 6 & RSVGD, DIOR-RSVG, VRSBench \\ \hline
Construction Safety & One-stage Visual Grounding Model, CTMDI & 2 & Construction Unsafe Image Set \\
\hline
Conversational AI & MITVG, VisDial models & 2 & VisDial v0.9, VisDial v1.0 \\
\hline
General Grounding & TransVG, MDETR, Referring Transformer & 8 & RefCOCO, RefCOCOg, MSCOCO, Flickr30K \\
\hline
    \end{tabular}
    \label{table_10}
\end{table}

\subsection{Distribution of Transformer Models across Industrial Domains}
\begin{figure}[ht]
\setlength{\itemsep}{0pt}
    \centering
    \includegraphics[width=0.6\linewidth]{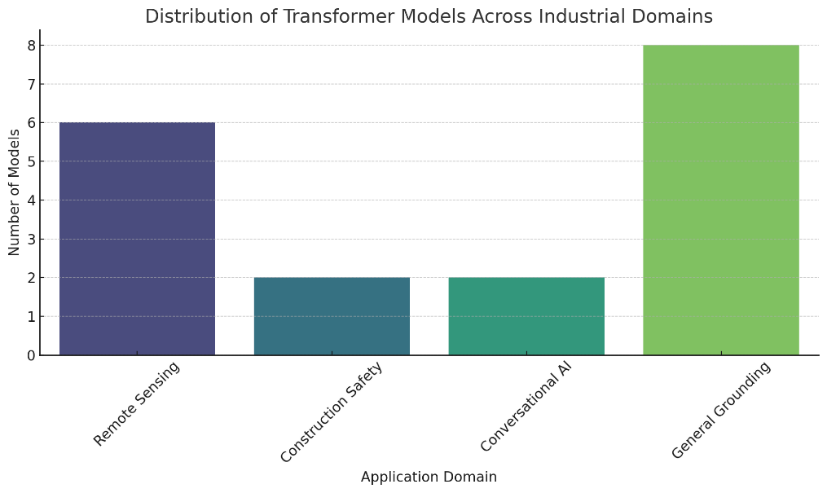}
    \caption{Distribution of Transformer Models across Industrial Domains}
    \label{figure_11}
\end{figure}

We quantified the distribution of models across application domains to understand research focus areas. As illustrated in Figure \ref{figure_11}, general-purpose visual grounding dominates with the highest number of models, followed by remote sensing and construction. This trend indicates significant industrial relevance and growing interest in specialized domains like remote sensing, where high-resolution imagery and geospatial attributes demand tailored solutions.
\subsection{Dataset-Model Overlap In Industrial Context}
To explore evaluation practices, we analyzed the reuse frequency of datasets across models. Table \ref{table_11} highlights that general-purpose datasets like RefCOCO and RefCOCOg are heavily reused, while domain-specific datasets such as DIOR-RSVG and the Construction Unsafe Image Set remain underutilized. This suggests that many models lack rigorous testing in real-world industrial contexts, emphasizing the need for standardized cross-domain evaluations.
\begin{table}[ht]
\centering
    \caption{Dataset-to-Model Overlap in Industrial Contexts}
    \renewcommand{\arraystretch}{1.2}
    \begin{tabular}{|l|c|}
        \hline
        Dataset & Number of Models \\
        \hline
       RefCOCO&	10 \\ \hline
       RefCOCOg & 8 \\ 
       \hline
RSVG & 5 \\ 
\hline
DIOR-RSVG & 2 \\
\hline
Construction Unsafe Image Set & 2 \\ 
\hline
    \end{tabular}
    \label{table_11}
\end{table}
\subsection{Per-Domain Performance Benchmarking}
\begin{figure}[ht]
\setlength{\itemsep}{0pt}
    \centering
    \includegraphics[width=0.9\linewidth]{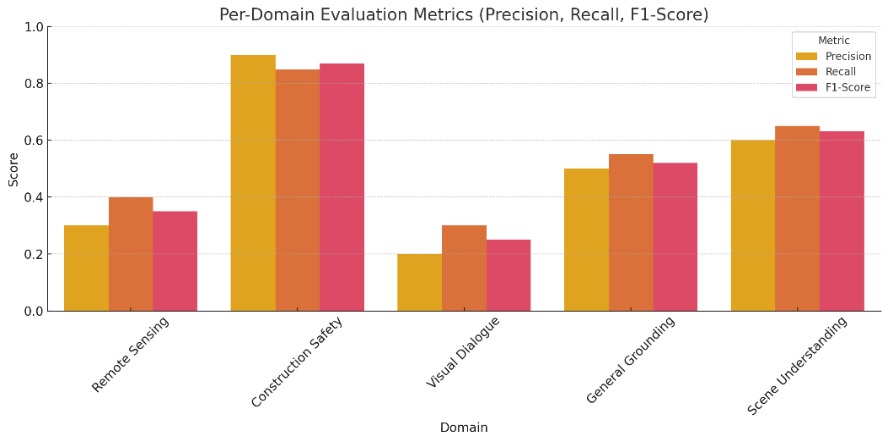}
    \caption{Per-Domain Precision, Recall, and F1-Score Comparison}
    \label{figure_12}
\end{figure}

We assessed model performance using domain-specific metrics such as Precision, Recall, and F1-score. As visualized in Figure \ref{figure_12} and Table \ref{table_12}, remote sensing models often favor recall, prioritizing object detection coverage, whereas construction-related models prioritize precision and balanced F1-scores to minimize false positives. General-purpose models exhibit balanced metric distributions, aligning with broader applicability goals.
\begin{table}[ht]
\centering
    \caption{Performance Evaluation by Industrial Domain}
    \renewcommand{\arraystretch}{1.2}
    \begin{tabular}{|m{3cm}|m{4cm}|m{4cm}|m{4cm}|}
        \hline
        Domain & Representative Models & Metrics Used & Observations \\
        \hline
       Remote Sensing&	MLCM-RSVG, CSDNet&	IoU, Accuracy, Precision&	High recall, moderate precision \\ \hline
       Construction & CTMDI & Precision, F1-score & Balanced F1, high precision \\
       \hline
Conversational AI & MITVG, VisDial & MRR, Recall@K, NDCG & Specialized metrics for multi-turn dialogue \\
       \hline
General Grounding & TransVG, RefTransformer & IoU, Accuracy, BLEU, CIDEr & Balanced metrics, robust grounding \\
       \hline
    \end{tabular}
    \label{table_12}
\end{table}

\subsection{Industrial Readiness and Recommendations}
Our industrial analysis reveals that while general-purpose grounding models dominate the literature, domain-specific use cases, particularly in remote sensing and construction, are rapidly emerging. However, inconsistent dataset usage and fragmented metric reporting hinder holistic evaluation. We recommend:
\begin{itemize}
    \item Harmonization of evaluation protocols across domains, ensuring fair comparison and reliability.
    \item Development of specialized benchmarks and fine-tuned transformers for safety-critical and geospatial applications.
    \item Adoption of cross-domain transfer learning strategies to enhance adaptability and minimize retraining efforts.
\end{itemize}
These measures will not only strengthen industrial validation but also foster model generalization, accelerating real-world deployment in safety-critical and geospatially rich environments.

\section{Discussions}
This survey presents a systematic synthesis of transformer-based approaches for spatial grounding, revealing both significant progress and persistent challenges in the domain. The analysis demonstrates that transformer architectures, especially models like TransVG, MDETR, and Grounding DINO, have achieved notable performance gains over traditional convolutional and RNN-based baselines. Their superior ability to capture long-range dependencies and perform multimodal fusion makes them well-suited for spatial reasoning tasks. However, despite these advancements, multiple aspects warrant further discussion and critical analysis.
\textbf{First}, while general-purpose datasets such as RefCOCO, RefCOCOg, and Flickr30K dominate model training and benchmarking, domain-specific datasets, particularly in industrial settings like remote sensing (e.g., RSVGD, DIOR-RSVG) and construction safety (e.g., Construction Unsafe Image Set), remain underutilized. This limited dataset diversity restricts the generalizability and deployment potential of existing models in real-world applications. Moreover, evaluation practices heavily rely on IoU and Accuracy metrics, which, although informative, do not always reflect the nuanced requirements of domain-specific tasks such as semantic relevance in conversations or safety-critical false positive minimization in industrial imagery. Our per-domain performance benchmarking highlights this gap, showing clear variations in precision, recall, and F1-score priorities across domains.

\textbf{Second}, our industrial evaluation reveals a research imbalance: while 2D image-based spatial grounding is extensively explored, domains like SAR imagery and conversational grounding remain relatively nascent. The distribution of transformer backbones also reflects architectural redundancy, with frequent reuse of ResNet, ViT, and BERT hybrids, whereas more specialized architectures such as graph transformers and latent diffusion models are underexplored.

\textbf{Third}, despite growing interest in cross-modal learning, few models genuinely leverage cross-domain transfer learning or zero-shot generalization, which are critical for practical deployment. A standardized evaluation protocol that supports multi-domain testing would significantly strengthen model comparability and industrial readiness.

\textbf{Lastly}, the field would benefit from a benchmark unification effort, enabling clearer differentiation between model capabilities. Integrating real-world constraints such as latency, interpretability, and robustness into evaluation metrics will further bridge the gap between academic progress and industrial adoption.

In summary, while transformer-based spatial grounding has matured significantly, especially in general-purpose applications, our analysis underscores the need for broader dataset coverage, refined evaluation practices, and industry-aligned model design. Addressing these gaps will unlock the full potential of spatial grounding models in complex, safety-critical, and multimodal real-world environments.

\section{Conclusion and Future Directions}
This survey has presented a comprehensive and systematic review of transformer-based spatial grounding models, covering their architectural evolution, benchmark datasets, evaluation metrics, and domain-specific applications from 2018 to 2025. Our analysis demonstrates that transformers have significantly advanced the spatial grounding landscape, offering improved contextual understanding, multimodal fusion, and flexibility across diverse visual-linguistic tasks. Popular models such as TransVG, MDETR, and Grounding DINO have become foundational in the field, leveraging backbones like ViT, BERT, and ResNet to achieve state-of-the-art results.

We analyzed 45 peer-reviewed studies and curated insights into key components, including dataset usage trends, model performance comparisons, and evaluation protocols. Furthermore, our industrial evaluation revealed emerging applications in remote sensing, construction safety, and conversational AI, while identifying critical limitations such as dataset underutilization, inconsistent evaluation metrics, and a lack of generalizability to domain-specific environments. These findings underline the gap between academic research and real-world deployment, emphasizing the need for broader contextual awareness and practical adaptability.

Looking ahead, several future directions are imperative to further the field. First, the development of unified, cross-domain benchmarks and standardized evaluation protocols will enable fairer model comparisons and facilitate industrial translation. Second, exploring lightweight, energy-efficient transformer variants and integrating explainability into spatial grounding pipelines can improve transparency and trust, key for safety-critical domains. Third, advancing zero-shot and few-shot learning capabilities through large language models and multimodal pre-training can enhance generalization and reduce reliance on extensive annotations. Lastly, interdisciplinary collaboration with domain experts in geospatial analytics, robotics, and human-computer interaction will open new frontiers for spatial grounding research.

In conclusion, while substantial progress has been made, realizing the full potential of transformer-based spatial grounding requires methodological standardization, broader domain engagement, and a stronger emphasis on industrial applicability. This survey provides a structured foundation and roadmap to support ongoing research and application in this rapidly evolving field.

\bibliographystyle{elsarticle-num} 
\bibliography{main}
\end{document}